\documentclass{article}
\usepackage[table, dvipsnames]{xcolor}
\usepackage[font=small, labelfont=bf]{caption}
\usepackage{multicol}
\usepackage[bookmarks=true, pagebackref]{hyperref}
\usepackage[square, numbers, sort&compress]{natbib}
\usepackage{lipsum}
\usepackage{xspace}
\usepackage{todonotes}
\usepackage{amsmath}
\usepackage{mathtools}
\usepackage{enumitem}
\usepackage{amssymb}
\usepackage{booktabs}
\usepackage[toc,page,header]{appendix}
\usepackage{minitoc}
\usepackage[ruled, vlined, linesnumbered]{algorithm2e}
\usepackage{wrapfig}
\usepackage{array}
\usepackage{xcolor, soul}
\usepackage{subcaption}
\usepackage{multirow}
\usepackage{fontawesome5}
\usepackage{pifont}
\usepackage{tabularx}
\usepackage[most]{tcolorbox}
\usepackage{algpseudocode}
\usepackage{cancel}
\usepackage{graphicx}
\usepackage{caption}
\usepackage{subcaption}
\usepackage[margin=1in]{geometry}
\usepackage[linesnumbered,ruled,vlined]{algorithm2e}
\usepackage{bm}
\usepackage{wrapfig}

\usepackage{etoolbox}

\usepackage[final]{corl_2025}

\begin{document}
% \maketitle

\title{Disentangled Multi-Context Meta-Learning:\\ Unlocking Robust and Generalized Task Learning}

\author{
  Seonsoo Kim\textsuperscript{*} \quad
  Jun-Gill Kang\textsuperscript{*} \quad
  Taehong Kim \quad
  Seongil Hong \\
  Agency for Defense Development \\
  \texttt{\{sunsoo3165,jungillkang,kimtaehong07,science4729\}@gmail.com}\\
  \vspace{1.0em}
  \url{seonsoo-p1.github.io/DMCM}
  \vspace{-2.0em}
}

\renewcommand{\thefootnote}{\fnsymbol{footnote}}
\footnotetext[1]{Equal contribution.}

\maketitle

%===============================================================================

\begin{abstract} 
    In meta-learning and its downstream tasks, many methods rely on implicit adaptation to task variations, where multiple factors are mixed together in a single entangled representation. This makes it difficult to interpret which factors drive performance and can hinder generalization. In this work, we introduce a disentangled multi-context meta-learning framework that explicitly assigns each task factor to a distinct context vector. By decoupling these variations, our approach improves robustness through deeper task understanding and enhances generalization by enabling context vector sharing across tasks with shared factors. We evaluate our approach in two domains. First, on a sinusoidal regression task, our model outperforms baselines on out-of-distribution tasks and generalizes to unseen sine functions by sharing context vectors associated with shared amplitudes or phase shifts. Second, in a quadruped robot locomotion task, we disentangle the robot-specific properties and the characteristics of the terrain in the robot dynamics model. By transferring disentangled context vectors acquired from the dynamics model into reinforcement learning, the resulting policy achieves improved robustness under out-of-distribution conditions, surpassing the baselines that rely on a single unified context. Furthermore, by effectively sharing context, our model enables successful sim-to-real policy transfer to challenging terrains with out-of-distribution robot-specific properties, using just 20 seconds of real data from flat terrain, a result not achievable with single-task adaptation.
\end{abstract}

% Two or three meaningful keywords should be added here

%===============================================================================
\begin{figure}[h!]
    \vspace{-1.3em}
    \centering
    \includegraphics[width=0.88\linewidth]{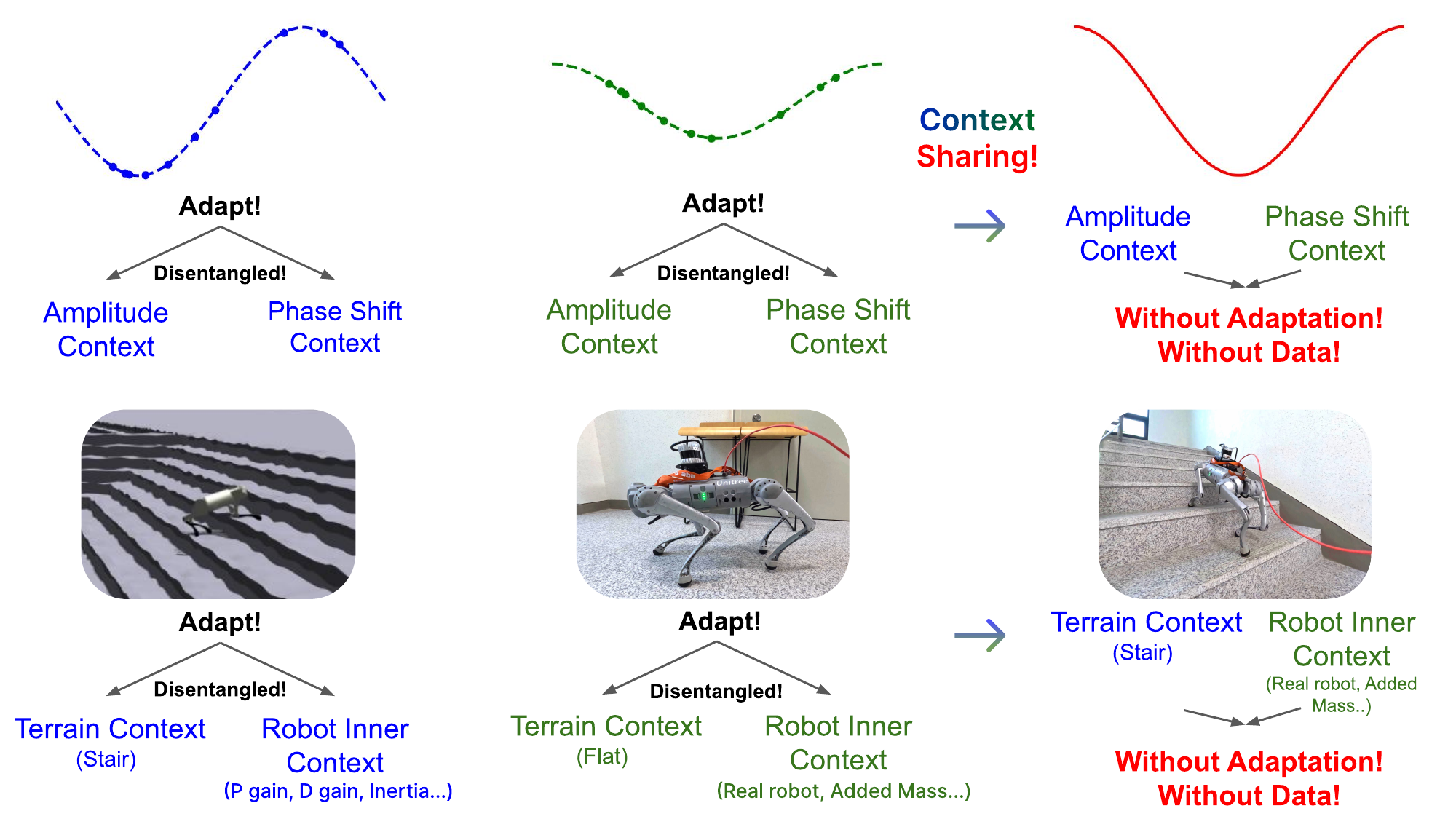}
    \vspace{-0.3em}
    \caption{Basic Concept of Disentangled Multi-Context Meta-Learning.
    Adaptation is illustrated for both the sine regression and robot dynamics tasks. The model adapts by disentangling task-specific factors into separate context vectors. These learned contexts can be reused across tasks with overlapping factors, enabling generalization. }
    \label{fig:one}
\end{figure}

\section{Introduction}
% \vspace{-0.3em}

Meta-learning, or ``learning to learn,'' was introduced as a general framework for task adaptation in machine learning \cite{LearningToLearn}. In robotics, where variations in physical conditions, environments, and dynamics can significantly affect task execution, meta-learning has emerged as a powerful approach for improving generalization and robustness~\cite{metarobotFin, kaushik2020fast, learningDynamic,neuralFly}.

% Gradient-based meta-learning methods like MAML typically learn a shared initialization or context parameters~\cite{MAML,leo,CAVIA,ANIL}  , but they often entangle task variations and assume all tasks are equally distinct. Structured approaches such as hierarchical meta-learning~\cite{hierarchical} address this partly, but still rely on a single latent representation.

Conventional Gradient-Based Meta-Learning (GBML) methods \cite{leo,CAVIA,ANIL,metasgd,reptile} including MAML~\cite{MAML} typically learn shared parameters for adapting to distinct task features. While effective, these approaches often assume all tasks are equally distinct, which leads to learning entangled representations of different variations. This entanglement can hinder the model’s transferability to novel situations~\cite{modular,CAVIA,pearl}, and may lead to conflicting gradients during learning~\cite{leo}. Therefore, for general and efficient adaptation, disentangling task variations is beneficial \cite{hierarchical,multimodal}. Such disentanglement has been explored not only in meta-learning, but also as a broader goal across diverse areas~\cite{disentangle,tripod,disentangle2,infogan,generalization,betavae,darla}, where building interpretable, disentangled, and general models is important.

To this end, we propose \textit{Disentangled Multi-Context Meta-Learning} (DMCM), a framework that learns multiple context vectors — each corresponding to a distinct factor of variation. For example, in sine regression, DMCM separates amplitude and phase contexts; in quadruped robot locomotion, it separates contexts for terrain characteristics and robot-specific properties. We build our method on CAVIA~\cite{CAVIA}, but unlike CAVIA, DMCM updates only the disentangled context vector relevant to the changed information.

This disentangled structure offers two key benefits: (1) \textbf{improved robustness}, by identifying which type of variation the model is encountering, and (2) \textbf{enhanced generalization}, by sharing context vectors across tasks that share factors, without adaptation. Fig.~\ref{fig:one} illustrates the main concept of our approach. DMCM is designed to capture distinct context vectors corresponding to different feature subsets within tasks. These context vectors can then be recombined and shared for adapting to novel tasks, where each partial context may not have been previously learned together.

We validate DMCM on sine regression and quadruped robot locomotion. In sine regression, DMCM outperforms MAML, CAVIA, and ANIL \cite{MAML,CAVIA,ANIL}, showing stable performance under out-of-distribution (OOD) conditions and enables zero-shot prediction via context sharing. We further analyze DMCM on the three-factor sine task with different numbers of contexts, comparing robustness and computational cost (Appendix~\ref{sec:Ncontext}). In quadrupedal robot locomotion, policies trained with DMCM-derived contexts outperform both CAVIA-based and vanilla policies under OOD conditions. Notably, DMCM achieves real-world stair climbing with OOD low $K_p$ gain through context sharing, even though the only real-world data used for adaptation was 20 seconds of flat-terrain walking.

Our contributions are two-fold.
\textbf{(1)} A meta-learning method that learns distinct factors of variation through regulated data sequencing.
\textbf{(2)} Demonstrating that disentangled contexts enable effective OOD adaptation and zero-shot transfer in sine regression and quadrupedal robot locomotion, including real-world deployment.

% \vspace{-2.0em}

%===============================================================================

%===============================================================================
\section{Preliminaries}
% \vspace{-0.3em}

\subsection{Gradient-Based Meta Learning}
% \vspace{-0.3em}

%Shorter Version 
Many GBML methods, such as MAML, follow a bi-level optimization scheme with an \textit{inner loop} for task-specific adaptation and an \textit{outer loop} for meta-optimization. Each task $\mathcal{T}_i$ has a dataset split into $\mathcal{D}_i^{\text{train}}$ and $\mathcal{D}_i^{\text{test}}$, used for inner-loop adaptation and outer-loop meta-update, respectively.

Let $\theta$ denote the shared parameters and $\phi$ the task-specific parameters. In CAVIA~\cite{CAVIA}, $\phi$ is a context parameter concatenated to the model, and only $\phi$ is updated during the inner loop, where $\phi_i$ is initialized to zero and $\alpha$ is the inner-loop learning rate:
% \vspace{-0.25em}
{\small
\begin{equation}
\phi_i \leftarrow \phi_i - \alpha \nabla_{\phi} \frac{1}{|\mathcal{D}_i^{\text{train}}|} \sum_{(x,y) \in \mathcal{D}_i^{\text{train}}} \mathcal{L}_{\mathcal{T}_i}(f_{\phi_i, \theta}(x), y)
\end{equation}}
\vspace{-0.75em}

The outer loop then updates $\theta$ using gradients from the test loss, where $\beta$ is the meta-learning rate and $N$ is the number of tasks per batch:
\vspace{-0.25em}

{\small
\begin{equation}
\theta \leftarrow \theta - \beta \nabla_{\theta} \frac{1}{N} \sum_{\mathcal{T}_i} \frac{1}{|\mathcal{D}_i^{\text{test}}|} \sum_{(x,y) \in \mathcal{D}_i^{\text{test}}} \mathcal{L}_{\mathcal{T}_i}(f_{\phi_i, \theta}(x), y)
\end{equation}}
\vspace{-0.5em}

Our approach extends CAVIA by introducing $K$ disentangled context vectors ${\phi^1, \dots, \phi^K}$, each representing a distinct task factor (e.g., amplitude, phase). These vectors are concatenated at the layers and initialized to zero. During learning, only the vector corresponding to the changed factor is updated, enabling selective, interpretable, and robust task adaptation.

\subsection{Reinforcement Learning for Quadrupedal Robot Locomotion}

We formulate the quadrupedal robot locomotion problem as a Partially Observable Markov Decision Process (POMDP) defined by the tuple $(\mathcal{S}, \mathcal{O}, \mathcal{A}, \mathcal{T}, r, \gamma)$, where $\mathcal{S}$, $\mathcal{O}$, and $\mathcal{A}$ denote the state, observation, and action spaces, respectively. $\mathcal{T}(s_{t+1} \mid s_t, a_t)$ defines the transition dynamics, $r(s_t, a_t)$ is the reward function, and $\gamma \in (0,1)$ is the discount factor ~\cite{pomdp}. Due to partial observability, the agent receives observations $o_t \in \mathcal{O}$ rather than full states $s_t \in \mathcal{S}$.

The objective is to learn a policy $\pi^*: \mathcal{O} \rightarrow \mathcal{A}$ that maximizes expected return:
\begin{equation}
\pi^* = \arg\max_{\pi} \, \mathbb{E}_{s_{t+1} \sim \mathcal{T},\, a_t \sim \pi(o_t)} \left[ \sum_{t=0}^{\infty} \gamma^t r(s_t, a_t) \right].
\end{equation}

To compensate for partial observability, many locomotion agents incorporate latent variables that capture task-relevant factors such as terrain properties and robot-specific properties~\cite{RMA,dreamwaq}. These latent representations are often learned via teacher-student frameworks or supervised loss signals. 
% Similar ideas have been explored in vision-based RL using disentangled representations~\cite{darla}, and in meta-reinforcement learning through inference of probabilistic latent contexts~\cite{pearl}, both of which highlight the importance of separating underlying factors for improved generalization and adaptation

While some prior works attempt to separate information using multiple encoders or decoders~\cite{loopsr,paloco}, our method explicitly disentangles multiple context vectors based on distinct task variations. This leads to more robust and interpretable adaptation in complex locomotion tasks.

\section{Disentangled Multi-Context Meta-Learning (DMCM)}

In this section, we provide a detailed description of DMCM, a framework that not only separates context vectors from model parameters but also disentangles the context vectors from each other. Each context vector is explicitly assigned to represent a distinct task factor, guided by the sequence of data provided during training. Pseudocode is shown in Alg.~\ref{alg:dmcm_2e}.

% \begin{figure}[h!]
%     \centering
%     \begin{subfigure}{0.3\textwidth}
%         \centering
%         \includegraphics[width=0.95\linewidth]{image/Fig2/sss.pdf}
%         \caption{Overview of DMCM explained with a diagram at $k$=2}
%         \label{fig:main}
%     \end{subfigure}
%     \hfill
%     \begin{subfigure}{0.665\textwidth}
%         \centering
%     \centering
%         \includegraphics[width=1.0\linewidth, trim=0 95 0 0]{image/Fig2/pse.pdf}
%         \caption{DMCM Pseudo code}
%         \label{fig:dmcm_algo}

%     \end{subfigure}
%     \caption{DMCM Algorithm with a brief diagram(a) and Pseudo code(b)}
%     \label{fig:fig2_pseudo_combined}
% \end{figure}

% \usepackage{amsmath,bm}
\SetKwInOut{Require}{Require} % Change Input to Require
\SetArgSty{textnormal}
\SetCommentSty{textnormal}
\SetAlFnt{\normalfont}

% Macro for cyclic subtraction on {1,...,k}
\newcommand{\cminus}[2]{\bigl((#1-1-#2)\bmod k\bigr)+1}

%Should add separtion loop part at appendix. just write simple here . 

\subsection{Context-Based Factor Separation}

DMCM requires an additional task-labeling step to associate each dataset with its underlying context factors. A task $\mathcal{T}_i$ is defined by $K$ context vectors, collectively represented as the task context $C_i$:
\begin{equation}
\mathcal{T}_i = \mathcal{T}_i(C_i), \quad \text{where } C_i = (c_i^1, \dots, c_i^K).
\end{equation}
For instance, in a sine regression task with varying amplitude and phase shift, $K=2$; $c^1$ represents amplitude and $c^2$ represents phase shift.

By explicitly separating these factors, DMCM learns specialized context vectors for each, enabling targeted adaptation and better generalization across combinations of factors.

\subsection{Inner Loop Training}

The inner loop update builds on CAVIA, but adapts only the relevant context vector. Let $s$ be the selected factor to be trained ($s \in \{1,2,\dots,K\}$). At each step, only the context vector $\phi_i^s$ corresponding to the changing factor $c^s$ is updated, while the others remain unchanged. Specifically, task ${C_i}$ is chosen to differ from ${C_{i-1}}$ only in $c^s$, with the other contexts identical.

% This targeted adaptation ensures that $\phi^s$ captures information specific to its corresponding context.

\begin{algorithm}[h!]
\setlength{\lineskip}{0pt}
\setlength{\lineskiplimit}{0pt}
\setlength{\itemsep}{0pt}
\setlength{\parskip}{0pt}

\footnotesize
\SetArgSty{textnormal}
\SetCommentSty{textnormal}
\SetAlFnt{\normalfont}
\caption{Disentangled Multi-Context Meta-Learning (DMCM)}
\label{alg:dmcm_2e}
\Require{Task distribution $p(\mathcal{T})$, conditional distribution $p(\mathcal{T}_i \mid \mathcal{T}_{i-1})$}
\Require{Inner learning rate $\alpha$, outer learning rate $\beta$, model $f_{\phi_0^1,\dots,\phi_0^K,\theta}$}
\Require{Warm-up task number $B$, Recombination enabled or not}

\textbf{Indexing:} For any $n$, $(s\pm n)$ is taken mod $K$ over $\{1,\dots,K\}$\;

\While{not done}{
    $\mathcal{L}_{\text{meta-basic}},\ \mathcal{L}_{\text{meta-recombination}} \gets 0$\;
  \For{each task $i = 0$ to $N-1$}{
    \eIf{$i = 0$}{
      Sample $\mathcal{T}_0 \sim p(\mathcal{T})$\;
    }{
      Sample $\mathcal{T}_i \sim p(\mathcal{T}_i \mid \mathcal{T}_{i-1})$\;
    }
    Split into $\mathcal{D}_i^{\text{train}}, \mathcal{D}_i^{\text{test}}$\;

    Initialize $\phi_i^s \gets 0$\;

    \For{inner-loop step}{
      $\phi_i^{s} \gets \phi_i^{s} - \alpha \nabla_{\phi^s}
      \frac{1}{|\mathcal{D}_i^{\text{train}}|}
      \sum\limits_{(x,y) \in \mathcal{D}_i^{\text{train}}}
      \mathcal{L}_{\mathcal{T}_i}\!\big(f_{\phi_i^1,\dots,\phi_i^K,\theta}(x), y\big)$\;
    }

    \If{$i \ge B$}{
      $\mathcal{L}_{\text{meta-basic}} \mathrel{+}= \frac{1}{N - B}\,
      \frac{1}{|\mathcal{D}_i^{\text{test}}|}
      \sum\limits_{(x,y) \in \mathcal{D}_i^{\text{test}}}
      \mathcal{L}_{\mathcal{T}_i}\!\big(f_{\phi_i^1,\dots,\phi_i^K,\theta}(x), y\big)$\;
    }
  
    \If{$i \ge B$ \textbf{and} \text{Recombination enabled}}{

      Temporarily load the saved $\bm{\bar{\phi}}
      = \left\{
        \phi_{i-(K+1)}^{\,s-1},\,
        \phi_{i-2(K+1)}^{\,s-2},\,
        \dots,\,
        \phi_{\,i-(K-1)(K+1)}^{\,s-K+1}
      \right\}$\;

      Sample $\mathcal{D}_j^{\text{test}}$ corresponding to $\bm{\bar{\phi}}$ and $\phi_i^{s}$\;

      $\mathcal{L}_{\text{meta-recombination}} \mathrel{+}= \frac{1}{N - B}\,
      \frac{1}{|\mathcal{D}_j^{\text{test}}|}
      \sum\limits_{(x,y) \in \mathcal{D}_j^{\text{test}}}
      \mathcal{L}_{\mathcal{T}_j}\!\big(f_{\phi_i^s, \bm{\bar{\phi}}, \theta}(x), y\big)$\;

      Save $\phi^{s+1}_{i}$ (i.e., $\phi_{i-K+1}^{\,s+1}$) for the later recombination loop\;

    }
        $s\mathrel{+}=1$
    
  }
  
  $\theta \gets \theta - \beta \nabla_\theta\!\big(\mathcal{L}_{\text{meta-basic}} + \mathcal{L}_{\text{meta-recombination}}\big)$\;

}
% \vspace{-1em}
\end{algorithm}

\subsection{Basic Outer Loop (Meta-gradient step)}

The basic outer loop updates the shared model parameters $\theta$ after completing $B$ warm-up tasks with their respective inner loops. This allows the context vectors $\{\phi^1, \dots, \phi^K\}$ to accumulate meaningful information before affecting $\theta$.

\begin{wrapfigure}[16]{r}{0.24\textwidth}
  \centering
  \vspace{-4em} % adjust if needed
  \includegraphics[width=\linewidth]{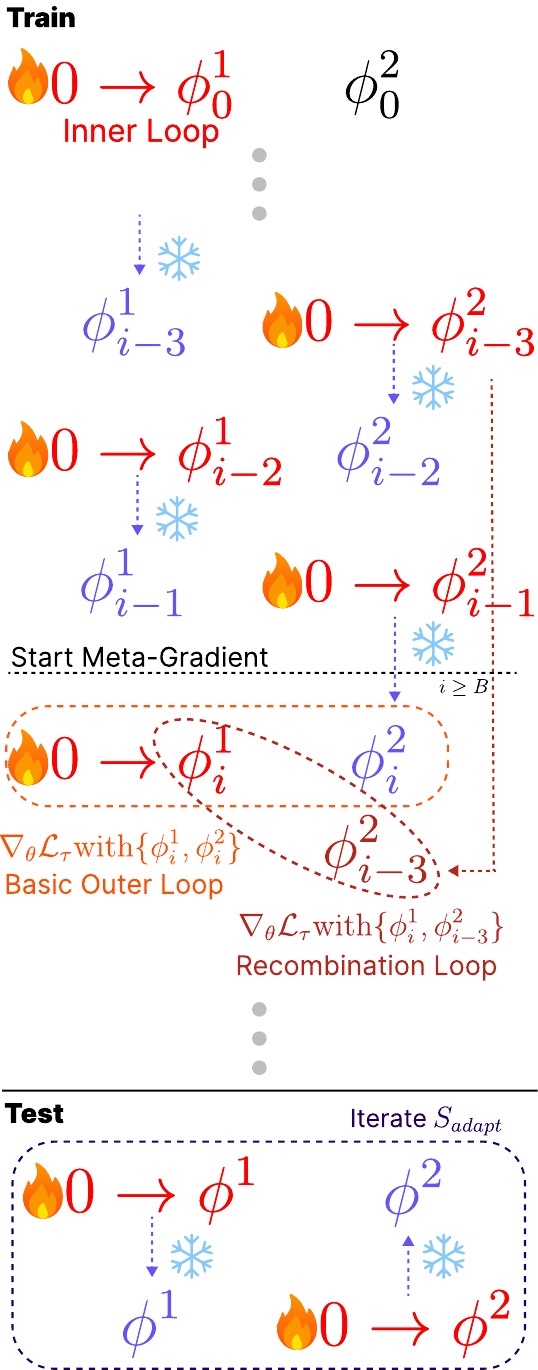}
  \caption{Simple diagram of DMCM for $K$=2 case}
  \label{fig:BasicDiagram}
  % \vspace{-4em}
\end{wrapfigure}
  % \vspace{-2em}

\subsection{Recombination Loop (Optional)}
\vspace{-0.5em}
To support zero-shot generalization when context vectors are shared, DMCM can optionally employ a \textit{recombination loop}.
In this procedure, the model parameters $\theta$ are updated using context vectors that were not adapted together during the inner loop.
This encourages the model to work effectively with independently adapted context vectors.
The detailed procedure and its effects are provided in Appendix~\ref{sec:NSeparation}.
\vspace{-0.5em}

% We collect two context vectors each from a different task adaptation and sample $D_j^{test}$ for the corresponding context. Then, update $\theta$ without additional inner loop adaptation.
% This forces the model to separate and rely on appropriate context information during adaptation, strengthening disentanglement. In this research, we only tried this at $K = 2$ case.

\subsection{Adaptation Procedure}
\vspace{-0.5em}

At test time, sequential adaptation is done from context $c^1$ to $c^K$ over $S_{adapt}$ iterations. This hyperparameter $S_{adapt}$ specifies the adaptation depth required for the task. If $S_{adapt}$ is too low, the model may fail to fully adapt to the target task, even if it has the capacity to do so.
\vspace{-0.5em}

\section{Experimental Results}

\vspace{-0.5em}

\subsection{Sine Task}

\vspace{-0.5em}

We evaluate DMCM’s effectiveness and the value of disentangled contexts using the standard sine regression benchmark~\cite{MAML}. Full experimental settings are provided in Appendix~\ref{sec:SineSetting}.
\vspace{-0.5em}

\subsubsection{Robustness under OOD Conditions}

\vspace{-0.5em}

To assess robustness, we compare DMCM with baselines—MAML, CAVIA, and ANIL~\cite{MAML, CAVIA, ANIL}—which are basic meta-learning algorithms known to work well at sine tasks.

We partition the amplitude and phase shift into five intervals, creating 25 range combinations. We then simulate out-of-distribution (OOD) conditions by excluding 10 combinations (40\% missing), 15 combinations (60\% missing), and 20 combinations (80\% missing) during training. 

Each method is evaluated under identical conditions (10 inner steps, same learning rate) for both 5-shot and 10-shot settings. 5-shot results are attached in Appendix~\ref{sec:5shot}.

As shown in Fig.~\ref{fig:sine_full}, CAVIA performs the best with full data, but its performance generally becomes unstable under exclusion of data as in Fig.~\ref{fig:sine40}. In contrast, DMCM achieves consistently lower loss and smaller variance, especially in high-exclusion settings. In Fig.~\ref{fig:20004000}, DMCM shows robustness for data exclusion at both 2000 and 4000 meta-gradient steps. This robustness stems from disentangling task factors, reducing sensitivity to missing combinations.

\begin{figure}[t]
    \centering
    \begin{minipage}[t]{0.49\textwidth}
        \centering
        \begin{subfigure}[t]{0.49\textwidth}
            \includegraphics[width=\linewidth]{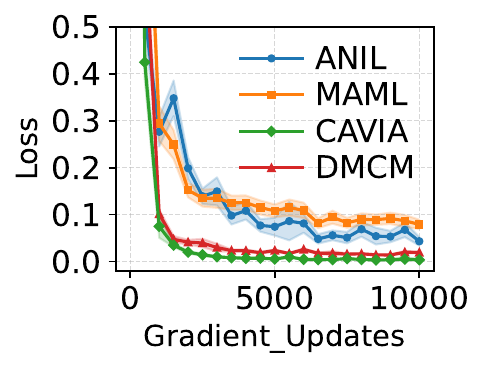}
            \caption{Full range (100\%)}
            \label{fig:sine_full}
        \end{subfigure}
        \hfill
        \begin{subfigure}[t]{0.49\textwidth}
            \includegraphics[width=\linewidth]{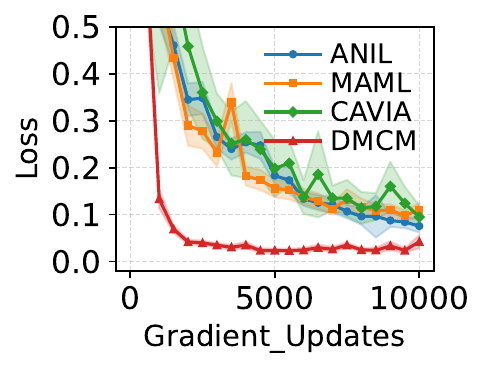}
            \caption{40\% Data range (60\% excluded)}
            \label{fig:sine40}
        \end{subfigure}
        \caption{
            10-shot sine regression with full data range and with 40\% data range used for training. Evaluation is performed on the full distribution. The shaded area shows a 95\% confidence interval.
            (a) 100\% of the range. (b) One case of 40\% range.
                    }
        \label{fig:totlasine}
    \end{minipage}
    \hfill
% --------- Figure 4 ---------
    \begin{minipage}[t]{0.49\textwidth}
        \centering
        \begin{subfigure}[t]{0.49\textwidth}
            \includegraphics[width=\linewidth]{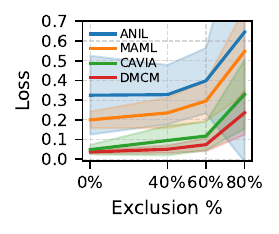}
            \caption{2000 Meta-gradients}
            \label{fig:2000}
        \end{subfigure}
        \hfill
        \begin{subfigure}[t]{0.49\textwidth}
            \includegraphics[width=\linewidth]{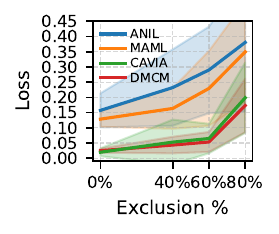}
            \caption{4000 Meta-gradients}
            \label{fig:4000}
        \end{subfigure}
        \caption{
            Average loss comparisons with range exclusion.
            (a) At 2000 meta gradients (near convergence for CAVIA/DMCM on full range), (b) At 4000 meta gradients (CAVIA/DMCM fully converged on full range).  
            The shaded region shows standard deviation. 30 randomly selected ranges were used for evaluation.
        }
        \label{fig:20004000}
    \end{minipage}

\vspace{-1.0em}

\end{figure}

\subsubsection{Zero-Shot Generalization via Recombination}

We assess DMCM’s effectiveness in assigning each factor of variation to the correct context vector and achieving zero-shot generalization to unseen tasks by sharing previously learned context vectors without further adaptation.

% Fig.~\ref{fig:separation_graph} shows training curves for self-adaptation loss and separation loss (zero-shot inference) using separated context vectors. 
\begin{wrapfigure}[16]{r}{0.63\textwidth} % height approx, adjust as needed
  \centering
  \vspace{-1.4em} % tune vertical alignment
  \begin{subfigure}[t]{0.49\linewidth}
      \includegraphics[width=\linewidth]{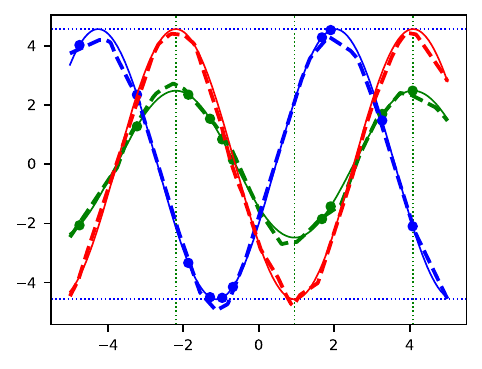}
      \subcaption{Two contexts: Amplitude (blue), Phase-Shift (green)}
      \label{fig:two_separation}
  \end{subfigure}
  \hfill
  \begin{subfigure}[t]{0.49\linewidth}
      \includegraphics[width=\linewidth]{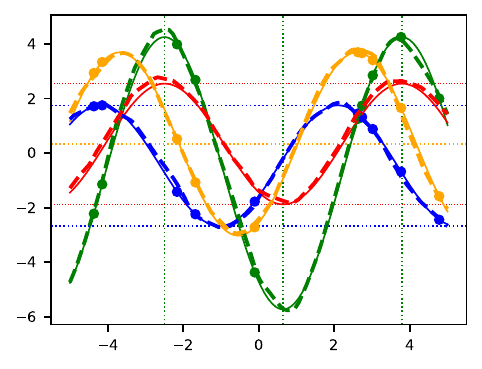}
      \subcaption{Three contexts: Amplitude (blue), Phase-Shift (green), Y-Shift (orange)}
      \label{fig:three_separation}
  \end{subfigure}
  \caption{Zero-shot prediction with disentangled context vectors. 
  Dotted lines (green, blue, orange) show predictions after adaptation to the displayed points, while the solid line denotes the ground truth. 
  Red dotted lines indicate predictions using only the shared context vectors \textbf{without adaptation}.}
  \label{fig:zero_shot_separation}
  \vspace{-2.0em}
\end{wrapfigure}

Fig.~\ref{fig:zero_shot_separation} visualizes predictions of unseen sine functions (red) using context vectors reused from other adaptations that share the same factors. This demonstrates DMCM’s ability to recombine contexts \textbf{without inner-loop adaptation}. Such capability not only enables zero-shot transfer but also allows selectively adapting or extracting specific context information, broadly applicable in diverse scenarios. Experimental settings are provided in Appendix~\ref{sec:zeroshot2} (two context) and ~\ref{sec:NSeparation} (three context).
 
 % Fig.~\ref{fig:three_separation} presents the three-context case, adding y-shift as a third factor. Detailed experimental settings are explained at Appendix~\ref{sec:zeroshot2} and additional results for various numbers of contexts are provided in Appendix~\ref{sec:Ncontext}.

\subsubsection{Number of Contexts for DMCM}
Tasks often involve multiple factors of variation, making the choice of context count critical for DMCM. Our experiments show that aligning the number of context vectors with the true factors of variation improves robustness, while using too many contexts slows adaptation. Thus, balancing robustness and efficiency is essential when deciding how many contexts to disentangle. Detailed results about the three-factor sine task, are provided in Appendix~\ref{sec:Ncontext}.

\subsection{Quadrupedal Robot Locomotion Task}
\label{sec:result}

\begin{figure}[t]
    \centering
    \includegraphics[width=1.0\linewidth]{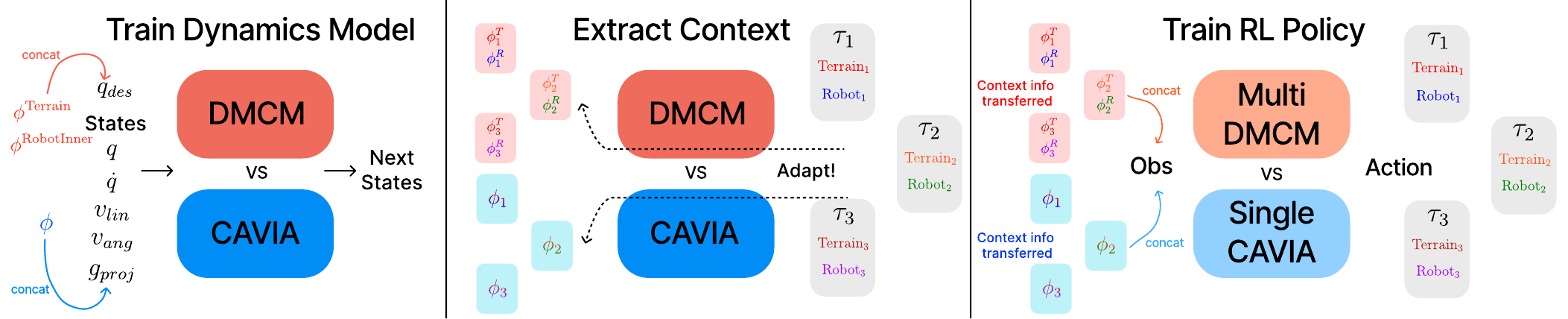}
    \caption{Learning procedure at Quadrupedal Robot Locomotion Task}
    \label{fig:robot_procedure}
    \vspace{-1em}
\end{figure}

% \vspace{-0.4em}

In this section, we first train the robot dynamics model using both CAVIA and DMCM, then extract their latent features—context parameters (CAVIA) and context vectors (DMCM)—across diverse simulation conditions. These extracted contexts are subsequently transferred as prior knowledge in RL, illustrated in Fig.~\ref{fig:robot_procedure}. We compare three RL policies under identical settings, differing only in the use of context information: a vanilla policy without context, a single-CAVIA policy using unified context parameters, and a multi-DMCM policy using disentangled context vectors (We set K=2).

% \vspace{-0.4em}

\subsubsection{Go1 Dynamics Model}

% \vspace{-0.4em}

We collect training data in simulation using a pretrained naive walking policy, whose performance is substantially worse than that of the baseline vanilla policy (Appendix~\ref{sec:naive}). This ensures that context extraction is not biased by high-performing behaviors. A dynamics model is then trained to predict joint positions ($q$), velocities ($\dot{q}$), projected gravity, linear velocities, and angular velocities from the robot state and desired joint positions ($q_{des}$) 0.02 seconds earlier.

\begin{figure}[b!]
    \vspace{-1em}
    
    \centering
    % --- First row ---
    \begin{subfigure}[t]{0.49\textwidth}
        \centering
        \includegraphics[width=\linewidth]{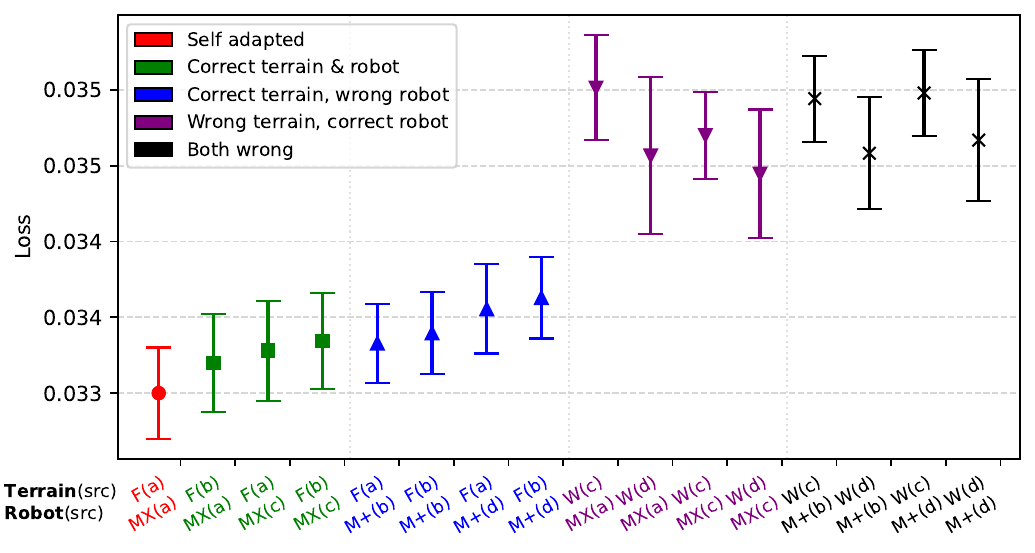}
        \vspace{-1.5em} % <-- vertical gap between rows

        \caption{Flat, MassX}
        \label{fig:wavy_mass_plus}
    \end{subfigure}
    \hfill
    \begin{subfigure}[t]{0.49\textwidth}
        \centering
        \includegraphics[width=\linewidth]{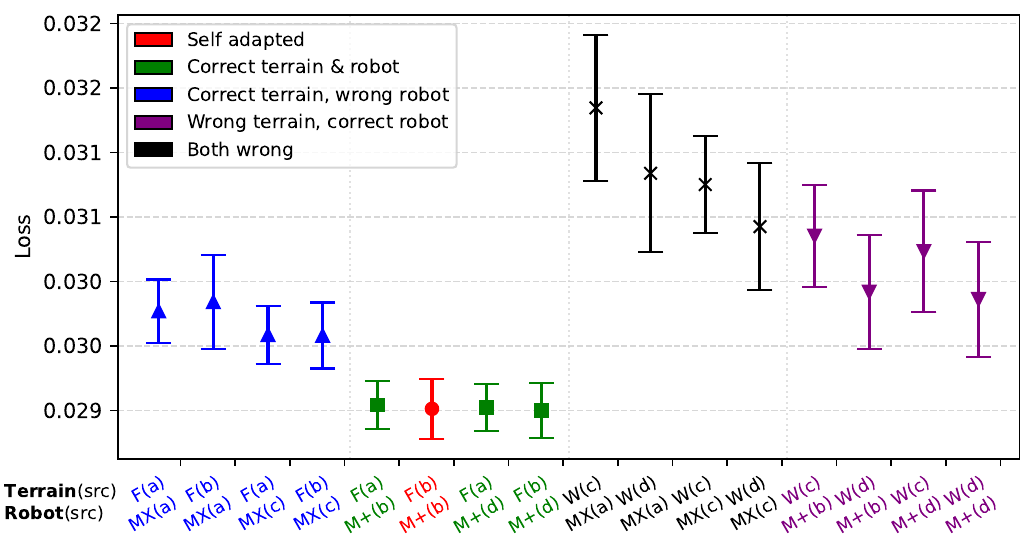}
        \vspace{-1.5em} % <-- vertical gap between rows

        \caption{Flat, Mass+}
        \label{fig:wavy_massx}
    \end{subfigure}
    
    \vspace{-0.3em} % <-- vertical gap between rows

    % --- Second row ---
    \begin{subfigure}[t]{0.49\textwidth}
        \centering
        \includegraphics[width=\linewidth]{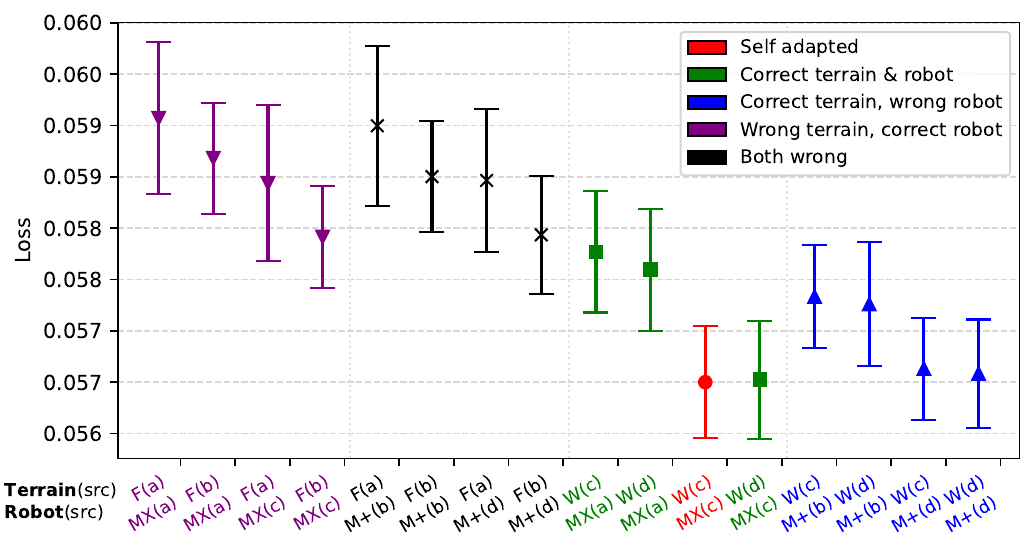}
    \vspace{-1.5em} % <-- vertical gap between rows

        \caption{Wavy, MassX}
        \label{fig:flat_mass_plus}
    \end{subfigure}
    \hfill
    \begin{subfigure}[t]{0.49\textwidth}
        \centering
        \includegraphics[width=\linewidth]{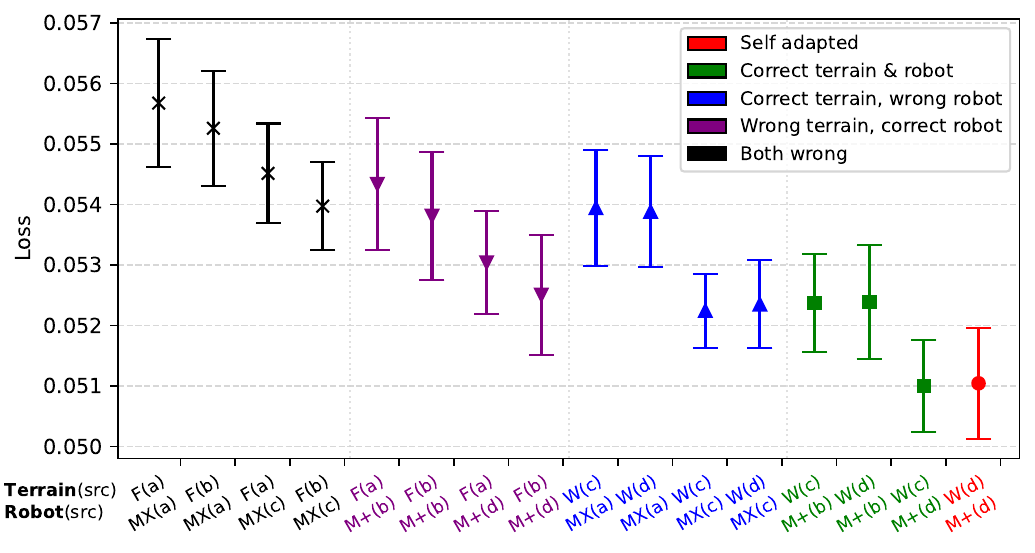}
    \vspace{-1.5em} % <-- vertical gap between rows

        \caption{Wavy, Mass+}
        \label{fig:flat_massx}
    \end{subfigure}
    \vspace{-0.5em} % <-- vertical gap between rows

    \caption{Each subplot shows the average loss for each dataset using 16 context vector sets derived from four real-world datasets. Each set includes a terrain (Flat (F) or Wavy (W)) and a robot property context (Additional Mass (M+) or No Additional Mass (MX)). Results are averaged over 10 trials, with adaptation and test sets randomly sampled from 20 seconds within the 40 seconds of data. Error bars indicate the standard deviation.}
    \label{fig:separation_eval}
    \vspace{-1em}

\end{figure}

Tasks are defined along two axes: \textit{terrain type} (44 variations, including flat, wavy, sloped, stair, and random terrains) and \textit{robot-specific properties} ($K_p$, $K_d$, payload, center of mass, delay, torque limits; full ranges in Appendix~\ref{sec:dynamicsHyper}). This yields $44 \times 350$ training tasks from Isaac Gym simulator~\cite{makoviychuk2021isaac}.

We train dynamics model using DMCM and CAVIA with matched hyperparameters and total context parameter counts, resulting in comparable prediction performance (see Appendix~\ref{sec:Go1Dynamics} for details).

% \vspace{-0.5em}

We evaluate the disentanglement capabilities of DMCM, trained only on simulation data, across four real-world datasets: (a) Flat terrain, (b) Flat terrain with 1.5 kg payload, (c) Wavy terrain, and (d) Wavy terrain with 1.5 kg payload. 
 Adaptation to these datasets yields terrain and robot-specific context vectors, allowing 16 possible context vector combinations for evaluation.

As shown in Fig.~\ref{fig:separation_eval}, models evaluated with context vectors aligned to the test data factor (red and green) generally achieve better performance than those using unrelated contexts. Notably, in some cases the shared context vector (green) even outperforms the self-adapted context vector (red). These results demonstrate that the disentanglement learned in simulation is transferred meaningfully to real-world dynamics.

%Sim2real 뺴도 되려나 한번 보고 결정. 
\vspace{-0.5em}
\subsubsection{Go1 Reinforcement Learning with Contexts}

We train RL policies with the extracted context parameters (CAVIA) and context vectors (DMCM), collected from all terrains and 500 robot-specific property settings, and concatenated them with observations during training. We evaluate the trained multi-DMCM, single-CAVIA and, vanilla policies in two complementary simulation experiments and real-world experiment.

\textbf{General OOD Evaluation} For evaluation, three environments are conducted: (1) The in-distribution terrain of steepest stair climbing, (2) OOD terrains with unseen height variations, (3) OOD robot-specific properties at the same terrain as (1).

For OOD robot property conditions, we apply a high payload (5–6 kg) and significantly reduced $K_p$ gains ($\Delta K_p$ between –5 and –4 $\text{N}\cdot\text{m}$), beyond the training ranges ($\leq 4$ kg payload and $\Delta K_p \geq -2.5 \,\text{N}\cdot\text{m}$). Full experimental details are provided in Appendix~\ref{sec:RLenvSetting} and~\ref{sec:eval_condition}

% \vspace{0.5em}

\begin{table}[h]
\vspace{-0.2em}

\centering
\begin{tabular}{lccc}
\hline
\textbf{Metric} & \textbf{Multi-DMCM} & \textbf{Single-CAVIA} & \textbf{Vanilla} \\
\hline
\multicolumn{4}{c}{\textbf{Inner Distribution}} \\
\hline

Success Counter & 1892/2000 & \textbf{1905/2000} & 1580/2000 \\
RMSE Linear Vel (m/s) & 0.0816 & \textbf{0.0798} & 0.1552 \\
Reward (w/o Termination Reward)& 21.9682 & \textbf{22.1789} & 17.4848 \\
Lifespan & 0.9851 & \textbf{0.9852} & 0.9190 \\
% AvgDist (All) & 7.8833 & 7.0405 & 6.5868 \\
AvgDist (Success)(m) & 8.4581 & \textbf{8.5203} & 7.3806 \\
\hline
\multicolumn{4}{c}{\textbf{OOD Terrain}} \\
\hline
Success Counter & \textbf{281/2000} & 149/2000 & 205/2000 \\
RMSE Linear Vel (m/s) & \textbf{0.2486} & 0.2844 & 0.2555 \\
Reward (w/o Termination Reward) & \textbf{10.0295} & 7.5442 & 8.5807 \\
Lifespan & \textbf{0.5476} & 0.4530 & 0.4737 \\

\hline
\multicolumn{4}{c}{\textbf{OOD Robot-Specific Properties}} \\
\hline
Success Counter & \textbf{602/2000} & 500/2000 & 48/2000 \\
RMSE Linear Vel (m/s) & \textbf{0.0944} & 0.1287 & \cancel{0.0826} \\
Reward (w/o Termination Reward) & \textbf{19.2283} & 17.8395 & \cancel{19.2308} \\
Lifespan & \textbf{0.6009} & 0.5844 & 0.3050 \\
% AvgDist (All) & 2.9941 & 3.4890 & 2.1075 \\
AvgDist (Success)(m) & \textbf{7.9205} & 6.0237 & 6.0990 \\
\hline
\end{tabular}

\caption{Comparison of three policies across inner distribution, OOD terrain, and OOD robot properties (based on 2000 trials) at simulation (Isaac Gym). Average distance is omitted for the OOD terrain due to the wide and variable range of linear velocity commands. For the OOD robot-specific properties case, the RMSE of linear velocity and reward values for the vanilla policy are not meaningful, as the robot mostly failed before climbing the stairs and only moved across flat regions.}
\label{tab:rl_eval}
\vspace{-0.8em}
\end{table}

As shown in Table~\ref{tab:rl_eval}, both the single-CAVIA and multi-DMCM policies outperform the vanilla baseline across all evaluation metrics under in-distribution conditions, showing that the context learned from the dynamics model is effective. Under OOD settings, the multi-DMCM model consistently achieves the best performance, thereby demonstrating robustness under unseen conditions.

% In contrast, the single-CAVIA model occasionally performs worse than the vanilla model despite achieving strong in-distribution performance. This result suggests that overfitting to a single, unified context parameter may harm generalization to unseen conditions, whereas disentangled multi-DMCM models retain better robustness. Additional simulations results comparing the models at gradually differing coditions are attached at Appendix~\ref{sec:AddSim}.

% \usepackage{graphicx} % in preamble

\textbf{Gradual Payload Variation} To assess robustness more systematically, we evaluate all policies in Isaac Gym on a 5 cm stair-climbing task under increasing payloads. As shown in Table~\ref{tab:mass_comparison}, all policies perform well under in-distribution payloads (0–3 kg). However, at OOD payloads (6 kg and 9 kg), the vanilla and single-CAVIA policies experience a sharp performance drop, while the multi-DMCM policy maintains strong performance, confirming its robustness to unseen physical variations.

\begin{table}[h!]
  \centering
  % \resizebox{0.7\textwidth}{!}{ % adjust 0.75 to control size
    \begin{tabular}{llcccc}
      \toprule
      \textbf{Metric} & \textbf{Method} & \textbf{0 kg} & \textbf{3 kg} & \textbf{6 kg (OOD)} & \textbf{9 kg (OOD)} \\
      \midrule
      \multirow{3}{*}{Lifespan}
        & Vanilla        & \textbf{0.998} & 0.996 & 0.986 & 0.897 \\
        & Single-CAVIA & 0.996 & 0.988 & 0.942 & 0.745 \\
        & Multi-DMCM   & 0.998 & \textbf{0.996} & \textbf{0.987} & \textbf{0.953} \\
      \midrule
      \multirow{3}{*}{Reward}
        & Vanilla        & 25.38 & 24.98 & 23.56 & 21.33 \\
        & Single-CAVIA & 24.60 & 24.32 & 23.00 & 19.61 \\
        & Multi-DMCM   & \textbf{25.51} & \textbf{25.09} & \textbf{24.27} & \textbf{22.62} \\
      \bottomrule
    \end{tabular}
  
  \caption{Performance comparison across gradual payload variations}
  \label{tab:mass_comparison}
  \vspace{-2em}
\end{table}

The single-CAVIA policy occasionally underperforms the vanilla policy in the two simulation experiments. This observation suggests that reliance on a single, unified context can induce overfitting and hinder robust application to unseen conditions, whereas disentangled multi-DMCM policy retains robustness. Additional simulation results of verifying context usage are attached at Appendix~\ref{sec:verify}.

\noindent \textbf{Real-world Deployment.}
Finally, we evaluate the RL policies on the Go1 robot in real-world conditions. All models perform reasonably when provided with appropriate context parameters or vectors. To further assess robustness, we deploy the three models under out-of-distribution (OOD) conditions by reducing the robot’s $K_p$ gains from 20 $\text{N}\cdot\text{m}$ to 16 $\text{N}\cdot\text{m}$ and adding a 1.5 kg payload while climbing 17 cm stairs, settings that were not encountered during training.

For evaluation, two pre-collected datasets are used as context: (i) simulation stair climbing data with 16 $\text{N}\cdot\text{m}$ $K_p$, and (ii) real flat-terrain data with 16 $\text{N}\cdot\text{m}$ $K_p$ and an additional payload. In the multi-DMCM policy, the terrain context was derived from (i) and the robot-specific context from (ii). In contrast, the single-CAVIA policy was evaluated separately with each dataset  because CAVIA cannot disentangle the two sources of variation.

\begin{table}[h!]
\centering
\small  % Reduce font size
\begin{tabular}{lcccc}
\toprule
\textbf{Metric} & \textbf{Multi-DMCM} & \textbf{Single (Sim Stair)} & \textbf{Single (Real Flat)} & \textbf{Vanilla} \\
\midrule
Success Rate (\%) & \textbf{80} & 0 & 0 & 40 \\
Avg. Time to Success (s) & \textbf{11.15} & NaN & NaN & 18.61 \\
\bottomrule

\end{tabular}
\caption{Success rate and average time to success in real-world OOD setting (5 trials, 17\,cm stairs, payload = 1.5 kg, $\Delta K_p = -4\, \text{N}\cdot\text{m}$).}
\label{tab:success_time}
\vspace{-1em}

\end{table}

\begin{wrapfigure}{r}{0.6\textwidth}
  \centering
  \vspace{-1em} % adjust vertical spacing if needed
  \includegraphics[width=\linewidth]{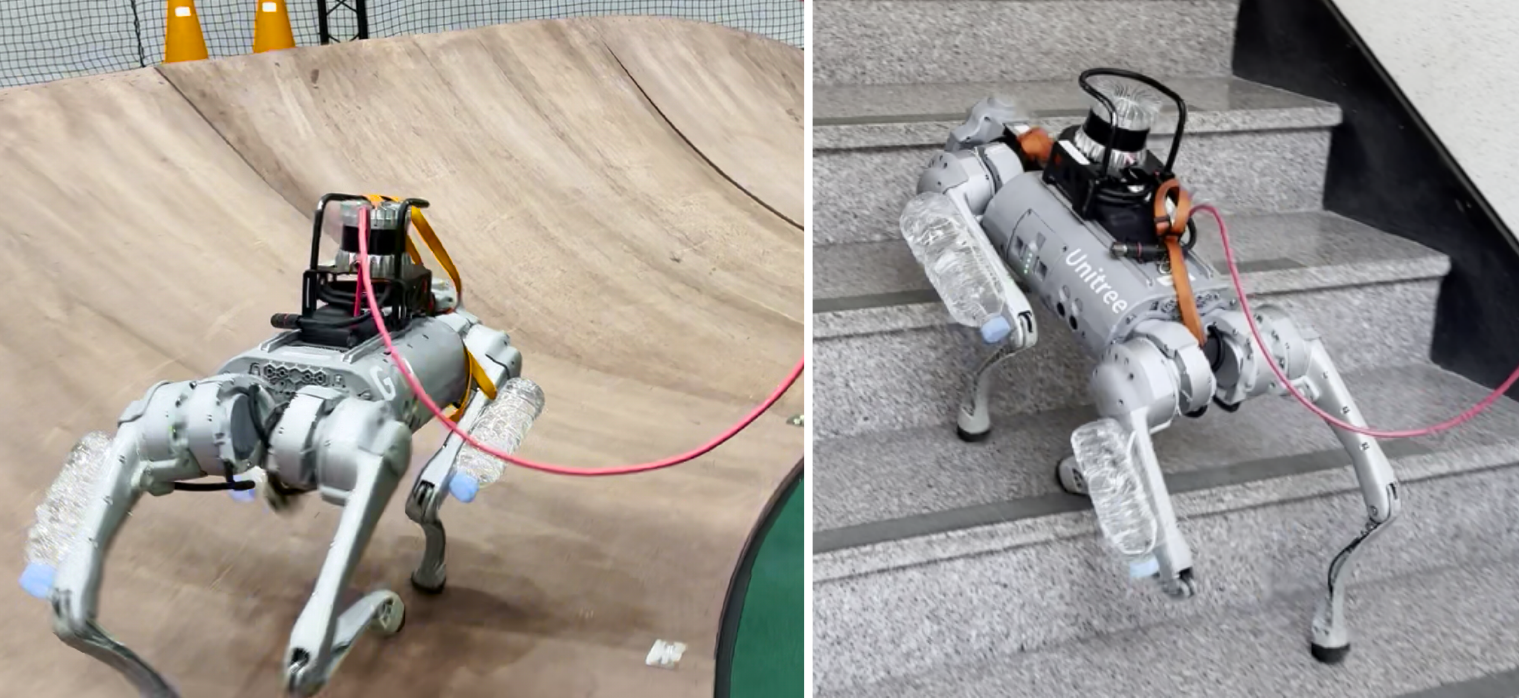}
  \caption{Additional deployment results showing the multi-DMCM policy on wavy terrain (left) and stair climbing (right) under asymmetric payloads with water bottles and a lidar sensor.}
  \label{fig:twopdfs}
\end{wrapfigure}

Using these contexts, the multi-DMCM policy achieves the best performance in climbing steep stairs at OOD conditions, leveraging its inherent robustness, whereas the single-CAVIA policy fails completely. Moreover, by sharing contexts, DMCM enables successful adaptation to stair conditions without any real-world stair data, relying only on 20 seconds of flat-terrain walking. Additional deployment results for multi-DMCM policy, along with detailed experimental settings, are provided in Appendix~\ref{sec:AddReal}.

%===============================================================================

\section{Conclusion}
We present Disentangled Multi-Context Meta-Learning (DMCM), a framework that learns multiple context vectors, each corresponding to a distinct factor of variation within tasks. Unlike prior approaches that rely on entangled or unified task representations, DMCM supports selective adaptation by updating only the relevant context vector. This design improves robustness by identifying which aspect of a task has changed, and enhances generalization by enabling context vector sharing across tasks with partially overlapping features.

Through experiments on sinusoidal regression and quadruped robot locomotion, we demonstrate DMCM’s robustness and generalization, including zero-shot adaptation to novel situations using prior experience.

Our work highlights the importance of disentangled context representations, offering a step toward interpretable, robust, and generalizable task adaptation in robotics and beyond.

\section{Limitations}

While DMCM improves robustness and generalization, several limitations remain:
\begin{itemize}
\item \textbf{Context labeling}: The current training process allows selecting the contexts to be learned, but it relies on manually defined, meaningful context labels with sufficient variation. Automating context discovery while preserving robustness and generalization is an important direction for future work.
\item \textbf{Task generality}: DMCM model itself has so far been validated only on regression tasks. Its extension to classification, reinforcement learning, and other domains remains to be explored.
\item \textbf{Real-time adaptation}: Although DMCM improves adaptability in quadruped robot locomotion tasks, our experiments relied on pre-collected data for adaptation. Incorporating meta-learning’s fast adaptation capability with selective context updates could enable robust real-time deployment in future settings.
\end{itemize}

Future work will therefore focus on automatic context discovery, broadening applicability across domains, and integrating selective real-time adaptation to further enhance DMCM’s utility in real-world robotics and beyond areas.

\section{Acknowledgments}
This work was supported by the Agency for Defense Development grant funded by the Korean
Government in 2025.

\label{sec:conclusion}

%===============================================================================

% \input{abstract_and_main}

\bibliography{ref}  % .bib
\clearpage  % or \newpage

\appendix
\part*{Appendix }

\setcounter{figure}{8}
\setcounter{table}{2}

\section{Sine Task}

\subsection{Hyperparameter and Experiment Settings}
To ensure a fair comparison across MAML, ANIL, CAVIA, and DMCM, we align the hyperparameters as closely as possible. All models use a meta learning rate of 0.001. The inner-loop learning rate is 0.1 for ANIL, CAVIA, and DMCM, but reduced to 0.001 for MAML, as higher values cause instability. Each algorithm applies 10 inner-loop updates.

Following common practice in sine regression tasks, the amplitude is sampled from $[0.1, 5.0]$ and the phase shift from $[0, \pi]$~\cite{CAVIA}. For the 10-shot case, 10 data points are provided for inner-loop training, meta-training, and evaluation, while in the 5-shot case, 5 data points are provided for each. For out-of-distribution evaluation, both the amplitude and phase shift ranges are uniformly divided into five intervals, resulting in a total of 25 range combinations for systematic exclusion.

All models employ a neural network with two hidden layers of 40 units each. Each meta-update uses a batch of 25 tasks. Evaluation is conducted on the same set of 500 sampled tasks, using mean squared error (MSE) computed over 100 test points.

For context dimensions, CAVIA uses 6 context parameters, while DMCM uses 3 parameters for each disentangled context vector. In DMCM, 10 warm-up iterations are performed before starting the outer loop ($B=10$), and 10 inner-loop iterations are applied for adaptation during evaluation ($S_{\text{adapt}} = 10$). Although all four algorithms use the same number of meta-gradient updates, DMCM applies additional inner-loop sequences before the warm-up stage. The recombination loop is not applied in DMCM.

\label{sec:SineSetting}

\subsection{5shot Result}

Same as 10 shot cases, we compare MAML, ANIL, CAVIA, and DMCM for 5shot cases. We sampled 30 different range selections same as 10 shot cases. 

\begin{figure}[h!]
    \centering
    \begin{subfigure}[t]{0.32\textwidth}
        \centering
        \includegraphics[width=\linewidth]{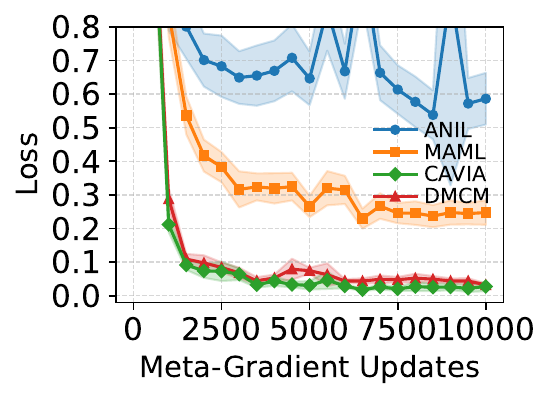}
        \caption{Full data at 5shot. Shaded ranges are 95\%confidence range}
        \label{fig:5shotfull}
    \end{subfigure}
    \hfill
    \begin{subfigure}[t]{0.32\textwidth}
        \centering
        \includegraphics[width=\linewidth]{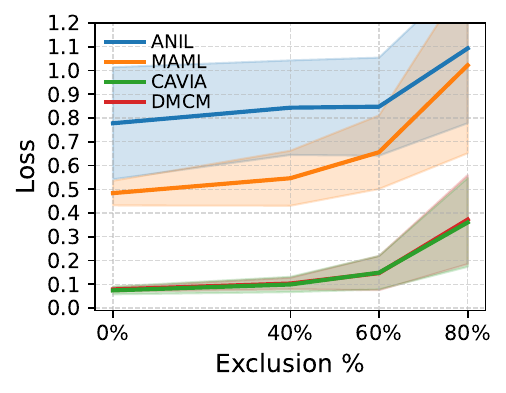}
        \caption{Average loss comparisons with range exclusions at 2000 meta-gradient steps. Shaded ranges are standard deviation.}
        \label{fig:5shot2000}
    \end{subfigure}
    \hfill
    \begin{subfigure}[t]{0.32\textwidth}
        \centering
        \includegraphics[width=\linewidth]{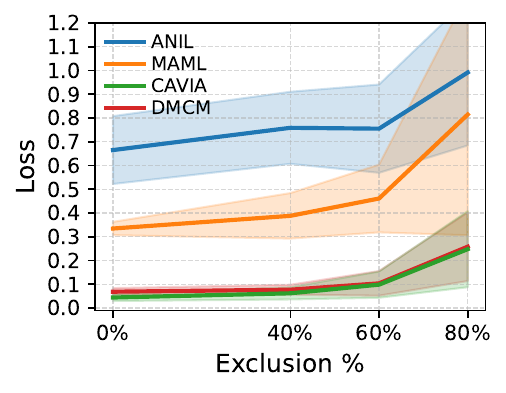}
        \caption{Average loss comparisons with range exclusions at 4000 meta-gradient steps. Shaded ranges are standard deviation.}
        \label{fig:5shot4000}
    \end{subfigure}
    \caption{Loss curves for (a) No exclusion, (b) average loss comparison with range exclusions at 2000 meta-gradients, and (c) 4000 meta-gradients}
    \label{fig:loss_curves}
\end{figure}

Unlike the results in the 10-shot cases shown in Fig.~3 and Fig.~4, DMCM and CAVIA exhibit similar performance under the same exclusion setting, as shown in Fig.\ref{fig:loss_curves}. We hypothesize that clear disentanglement requires sufficient data to capture the underlying task factors. Nevertheless, both CAVIA and DMCM outperform the other methods.

% \begin{figure}[t]
%     \centering
%     \subfigure[2000 meta-gradients]{
%         \includegraphics[width=0.45\textwidth]{image/Fig3/5shot2000_Sup.pdf}
%         \label{fig:left}
%     }\hfill
%     \subfigure[4000 meta-gradients]{
%         \includegraphics[width=0.45\textwidth]{image/Fig3/5shot4000_Sup.pdf}
%         \label{fig:right}
%     }
%     \caption{Overall figure caption}
%     \label{fig:twopdfs}
% \end{figure}
\label{sec:5shot}

\subsection{Analysis of Zero Shot Prediction}

\label{sec:zeroshot2}

As illustrated in Fig.~5, the trained DMCM model successfully predicts sine functions without direct access to task data, relying solely on shared context vectors. The model uses 3 parameters for each context vector, an inner-loop learning rate of 0.08 with a decay factor of 0.92 applied at each step, and an outer-loop learning rate of 0.001. It performs 30 inner-loop steps, begins meta-gradient updates after 10 tasks ($B=10$), and employs a neural network with two hidden layers of 40 units each. Each meta-update aggregates over 25 tasks, and evaluation is conducted using a 3-step iterative adaptation ($S_{\text{adapt}}=3$). Two context vectors are concatenated with the input. In addition, the recombination loop is applied.

\begin{figure}[h!]
    \centering
    \vspace{-1em}
    \begin{minipage}{0.45\textwidth}
        \centering
        \includegraphics[width=\linewidth]{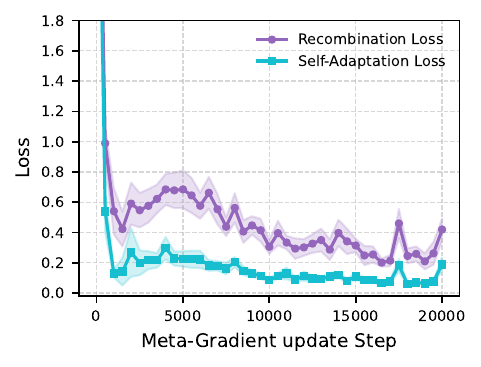}
        \caption{Learning of Sine Task with DMCM with recombination loop. }
        \label{fig:separation_graph}
    \end{minipage}
    \hfill
    \begin{minipage}{0.45\textwidth}
    
        \includegraphics[width=\linewidth]{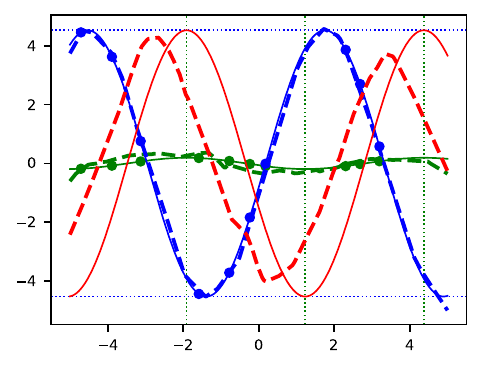}
        \caption{Example of an incorrect zero-shot prediction when the phase-shift context is derived from a low-amplitude case (green). The zero-shot prediction (red dot) deviates significantly from the ground truth (red solid line) }
        \label{fig:lowAmp}
    \end{minipage}
    \vspace{-1em}

\end{figure}

In our experiments, recombination loss, reflecting zero-shot prediction using shared contexts, is generally higher than self-adaptation loss, which involves direct adaptation using task specific data. This trend holds in sine regression as well, shown at Fig.~\ref{fig:separation_graph}. However, we find that the important contributor to the higher recombination loss is the difficulty of extracting meaningful phase context from low-amplitude sine waves.

% As shown at Fig.~\ref{fig:lowAmp}, the DMCM model struggles to infer phase shift when the amplitude is very low. To analyze this further, we compare three cases:  (i) self-adaptation loss, (ii) separation (zero-shot) loss over the full amplitude range, and (iii) separation loss restricted to the amplitude range $[1.5, 5.0]$.
% \begin{table}[h!]
% \centering
% \begin{tabular}{lcc}
% \toprule
% \textbf{Method} & \textbf{Mean Loss} & \textbf{95\% Confidence} \\
% \midrule

% Self Adapted & 0.0578 & $\pm$ 0.0051 \\
% Separation (Zero-Shot, Full Range) & 0.1876 & $\pm$ 0.0416 \\
% Separation (Zero-Shot, Amp [1.5, 5.0]) & 0.0893 & $\pm$ 0.0090 \\

% \bottomrule
% \end{tabular}
% \caption{Mean loss and 95\% confidence intervals. 500 tasks are randomly sampled for evaluation.}
% \label{tab:adaptation_loss}
% \end{table}

% As Table~\ref{tab:adaptation_loss} shows, high separation loss was mainly derived from shared context vector that does not include enough information. This result shows that zero-shot prediction becomes difficult if shared context vector is derived from data who does not contain enough information. 

As shown in Fig.~\ref{fig:lowAmp}, the DMCM model struggles to infer the phase shift context when the amplitude is very low. To examine this effect, we compare three cases: (i) self-adaptation loss, (ii) recombination (zero-shot) loss over the full amplitude range, and (iii) recombination loss restricted to the amplitude range $[1.5, 5.0]$.

\begin{table}[h!]
    \vspace{-1em}

\centering
\begin{tabular}{lcc}
\toprule
\textbf{Method} & \textbf{Mean Loss} & \textbf{95\% Confidence} \\
\midrule

Self Adapted & 0.0578 & $\pm$ 0.0051 \\
Recombination (Zero-Shot, Full Range) & 0.1876 & $\pm$ 0.0416 \\
Recombination (Zero-Shot, Amp [1.5, 5.0]) & 0.0893 & $\pm$ 0.0090 \\

\bottomrule
\end{tabular}
\caption{Mean loss and 95\% confidence intervals. 500 tasks are randomly sampled for evaluation.}
\label{tab:adaptation_loss}
\end{table}

\vspace{-1.0em}

As shown in Table~\ref{tab:adaptation_loss}, the recombination loss is significantly reduced—from 0.1876 to 0.0893—when restricting the amplitude range to $[1.5, 5.0]$. This $52.4\%$ reduction suggests that high recombination loss in the full-range case is primarily due to the shared context vector being derived from inputs with insufficient information (i.e., very low amplitudes). These results support the hypothesis that zero-shot prediction becomes unreliable when the context vector is formed from data that lacks enough information to represent the underlying task.

\vspace{-0.5em}

\begin{figure}[h!]
    \centering

    \begin{subfigure}[t]{0.23\textwidth}
        \centering
        \includegraphics[width=\linewidth]{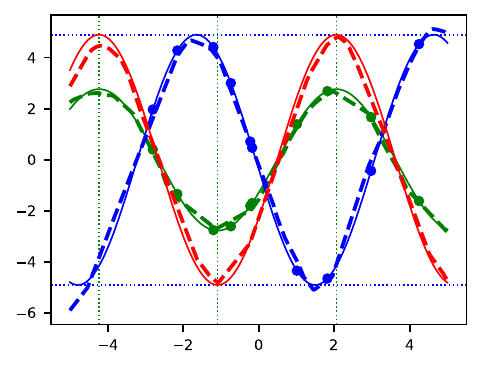}
        \label{fig:plot1}
    \end{subfigure}
    \hfill
    \begin{subfigure}[t]{0.23\textwidth}
        \centering
        \includegraphics[width=\linewidth]{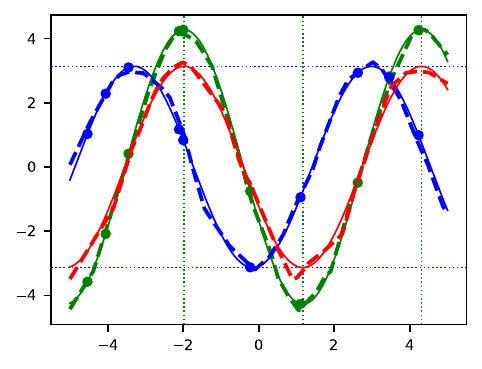}
        \label{fig:plot2}
    \end{subfigure}
    \hfill
    \begin{subfigure}[t]{0.23\textwidth}
        \centering
        \includegraphics[width=\linewidth]{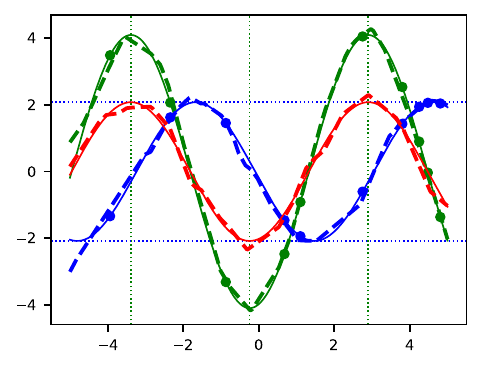}
        \label{fig:plot3}
    \end{subfigure}
    \hfill
    \begin{subfigure}[t]{0.23\textwidth}
        \centering
        \includegraphics[width=\linewidth]{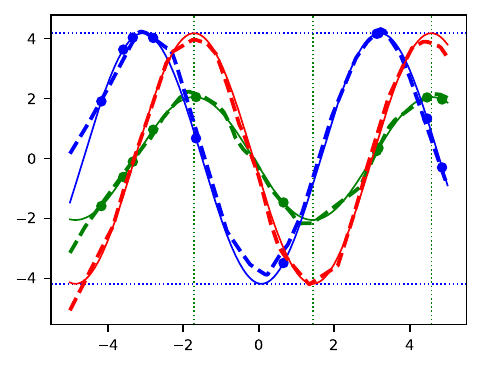}
        \label{fig:plot4}
    \end{subfigure}

    \caption{Additional zero-shot results from DMCM with two contexts. 
    Zero-shot predictions (red dots) are obtained through shared context vectors. 
    Tasks used for adapting amplitude and phase-shift contexts are shown in blue and green, respectively.}
    \label{fig:loss_grid}
\end{figure}

\subsection{Mislabeled Contexts}
We evaluate the effect of mislabeled task data on the sine regression task. Hyperparameters and training settings follow Appendix~\ref{sec:SineSetting}, and evaluation is conducted after 6000 meta-gradient steps. As shown in Table~\ref{tab:mislabeled}, both DMCM and CAVIA show similar degradation as the proportion of mislabeled data increases. Here, “10\% mislabeling” means that during task sampling, each task has a 10\% probability of being assigned an incorrect task label.

\begin{table}[h!]
\centering
\small
\begin{tabular}{lcccc}
\toprule
\textbf{Model} & \textbf{Clean} & \textbf{10\% mislabeled} & \textbf{20\% mislabeled} & \textbf{Increase (\%)} \\
\midrule
DMCM  & 0.0225 & 0.0476 & 0.0631 & +111\% / +180\% \\
CAVIA & 0.0135 & 0.0267 & 0.0467 & +98\% / +246\% \\
\bottomrule
\end{tabular}
\caption{Effect of mislabeled context data on sine regression tasks. Percentage increase is relative to the clean baseline.}
\label{tab:mislabeled}
\end{table}

\section{Additional Analysis of DMCM: N Context Sine}
\label{sec:Ncontext}
To explore DMCM’s scalability to multiple task factors and further analysis, we extend our evaluation to a setting with three factors of variation:
(i) amplitude,
(ii) phase shift, and
 (iii) y-shift(vertical offset).

Each sine function is modeled as:

\begin{equation}
y = A \cdot \sin(x - \phi) + b,
\end{equation}

where Amplitude $A \in [0.1,\ 5.0]$, Phase shift $\phi \in [0,\ \pi]$, Y-shift $b \in [-2.0,\ 2.0]$.

\subsection{Robustness with Varying Numbers of Contexts}
We compare models with 1 to 4 context vectors on sine regression tasks involving amplitude, phase shift, and y-shift variations.  
For the 2-context case, we test (i) a variant with amplitude and y-shift only (phase missing), and (ii) a variant combining amplitude and y-shift in one context and phase shift in another.
For the 4-context setting, we introduce redundancy by assigning two separate context vectors to the amplitude term, resulting in
$(A_1 \cdot \sin(x - \phi) + A_2 \cdot \sin(x - \phi) + b)$.

\label{sec:NcontextParam}

\paragraph{Hyperparameters and Experimental Settings.}
Models are evaluated under full-range and exclusion-range training. All experiments use an inner-loop learning rate of $0.05$ with a decay factor of $0.92$ per step, and an outer-loop (meta) learning rate of $0.00033$.
Each model is trained for $20$ inner-loop steps using a fully connected network with four hidden layers of $40$ units each.
We sample $45$ tasks per meta-update. For DMCM, meta-gradient updates begin after $B=20$ warm-up tasks.
During evaluation, we use $S_{\text{adapt}} = 5$ iterative loops for the full-range setting and $S_{\text{adapt}} = 10$ loops for the OOD setting. Recombination loop is not applied.

\paragraph{Context Configurations.}
We evaluate the following context configurations:
\begin{itemize}
    \item \textbf{1[A,P,Y] (CAVIA)}: A single context encodes amplitude, phase, and y-shift jointly (9 parameters), concatenated with input. 
    \item \textbf{2[A][Y] (Phase shift missing)}: Two contexts (4 parameters each), encoding amplitude and y-shift separately, without phase-shift context. Both concatenated with input.
    \item \textbf{2[A,Y][P]}: Two contexts (4 parameters each), where one encodes amplitude and y-shift jointly, and the other encodes phase shift. Both concatenated with input.
    \item \textbf{3[A][Y][P]}: Three contexts (3 parameters each), encoding amplitude, y-shift, and phase shift separately. Amplitude context is concatenated at the first layer, y-shift at the second layer, and phase shift with input.
    \item \textbf{4[A1][A2][Y][P]}: Four contexts (2 parameters each), where amplitude is split into two contexts ($A_1$, $A_2$) with ranges $[0.05, 2.5]$, plus y-shift and phase-shift contexts. $A_1$ and $A_2$ are concatenated at the first layer, y-shift at the second layer, and phase shift with input.
\end{itemize}

\begin{figure}[htbp]
    \centering
    \begin{subfigure}[t]{0.49\linewidth}
        \centering
        \includegraphics[width=\linewidth]{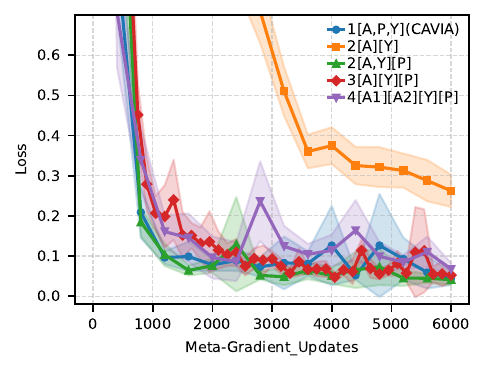}
        \caption{Full-range training and evaluation.}
        \label{fig::sine_full}
    \end{subfigure}
    \hfill
    \begin{subfigure}[t]{0.49\linewidth}
        \centering
        \includegraphics[width=\linewidth]{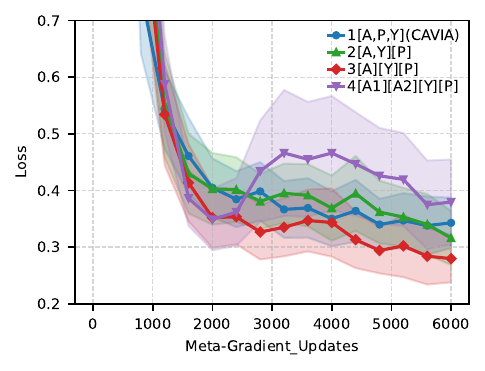}
        \caption{OOD evaluation over 40 training range samples.}
        \label{fig::sine_exclusion}
    \end{subfigure}
    \caption{Training curves of N-context models under (a) full-range and (b) OOD evaluation. 
Shaded areas represent confidence intervals: 95\% for (a) and 0.25 standard deviation for (b). 
The two-context case without the phase-shift factor is excluded in (b) due to extremely poor performance.}
    \vspace{-1.5em}
    
\end{figure}

In the full-range setting, all models perform well unless a critical variation is excluded from the contexts (the two-context model without phase shift).  
For OOD evaluation, each factor range is divided into four intervals, with only two intervals used for training. This corresponds to 12.5\% of the total range being covered during training, except in the four-context case (where each amplitude subrange $[0.05, 2.5]$ is divided and sampled separately).  
Under these OOD conditions, DMCM with the exact number of variation-specific contexts (3) achieves the best performance, consistently yielding low loss.  
By contrast, adding redundant context makes learning less stable and increases variability during training, as seen in the 4-context model. 

% \vspace{2em}
%Space for later change.. 

\vspace{-0.7em}
\subsection{N-Context Recombination}
\vspace{-0.3em}

Although Fig.~\ref{fig:BasicDiagram} illustrates only the $K=2$ case, our method naturally extends to $N$ contexts.  
The core idea of the recombination loop is to update the model using context vectors that were not jointly adapted in the inner loop.  
This is achieved by saving $K-1$ context vectors for each of the $K$ contexts and reusing them during training.

For example, consider an adaptation sequence of \{y-shift, amplitude, phase shift\}.  
When updating the y-shift context at iteration $i$, we require:
\begin{itemize}
\vspace{-0.5em}

    \item the phase-shift context vector adapted at $i - (K+1) = i - 4$, and
    \item the amplitude context vector adapted at $i - 2(K+1) = i - 8$,
    \vspace{-0.5em}

\end{itemize}
ensuring that none of these were adapted together in the same inner loop.

This mechanism requires $(K-1)\times K$ additional context vectors in memory; for $K=3$, this corresponds to six extra context vectors.    
Without the recombination loop, zero-shot prediction fails; in our experiments, the recombination loss increased from $0.069$ (with loop, restricted amplitude range) to $3.66$ (without loop, restricted amplitude range).
While DMCM still provides robustness without the recombination loop, zero-shot recombination is only possible when the loop is applied.
However, introducing it slightly reduces self-adaptation performance.  
Learning settings follow Appendix~\ref{sec:NcontextParam}, with recombination loop and amplitude restrictions as in Appendix~\ref{sec:zeroshot2}.
\label{sec:NSeparation}

\begin{figure}[h!]
    \centering

    \begin{subfigure}[t]{0.23\textwidth}
        \centering
        \includegraphics[width=\linewidth]{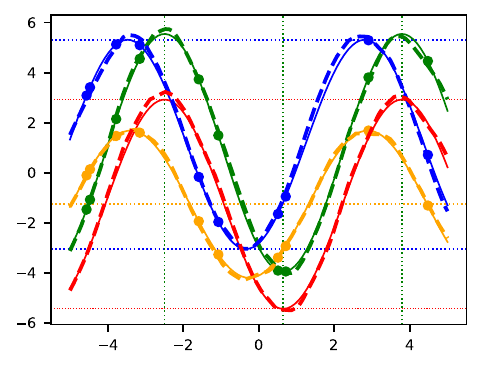}
        \label{fig:three1}
    \end{subfigure}
    \hfill
    \begin{subfigure}[t]{0.23\textwidth}
        \centering
        \includegraphics[width=\linewidth]{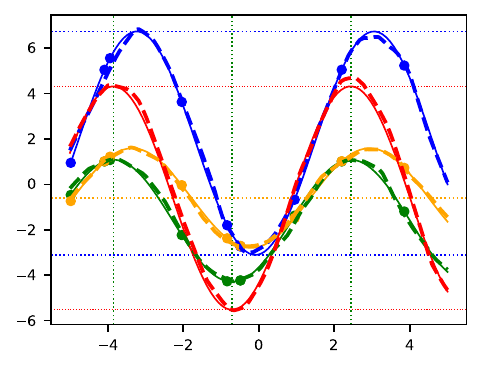}
        \label{fig:three2}
    \end{subfigure}
    \hfill
    \begin{subfigure}[t]{0.23\textwidth}
        \centering
        \includegraphics[width=\linewidth]{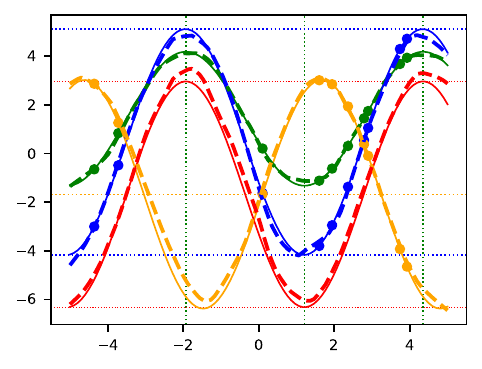}
        \label{fig:three3}
    \end{subfigure}
    \hfill
    \begin{subfigure}[t]{0.23\textwidth}
        \centering
        \includegraphics[width=\linewidth]{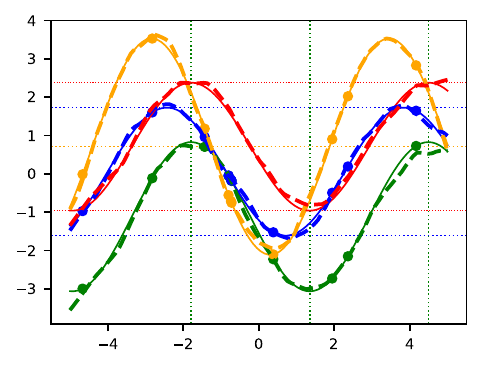}
        \label{fig:three4}
    \end{subfigure}
    \caption{Additional zero-shot results from DMCM with three contexts. 
Zero-shot predictions (red dots) are obtained through shared context vectors. 
Tasks used for adapting amplitude, phase-shift, and y-shift contexts are shown in blue, green, and orange, respectively.}
    \label{fig:loss_grid}
\end{figure}

\subsection{Computation of N Context Cases}
In Table~\ref{tab:training_adaptation_time}, we compare the training time of MAML, CAVIA, and DMCM on the sine task with varying numbers of context vectors listed at Appendix~\ref{sec:NcontextParam}. For DMCM, training time per meta-gradient decreases as the number of context vectors increases, due to smaller update sizes from fewer parameters per vector. However, adaptation time grows with more contexts, since DMCM adapts each context separately and requires $S_{\text{adapt}}$ iterations for all contexts to interact sufficiently. In this work, we set $S_{\text{adapt}} = 5$, though this value may vary depending on the task.
\label{sec:Computation}

\begin{table}[h!]
\scriptsize
\setlength{\tabcolsep}{3pt} % default is 6pt
\centering
\begin{tabular}{lccccc}
\toprule
\textbf{} & \textbf{MAML} & \textbf{CAVIA (1)} & \textbf{DMCM (2)} & \textbf{DMCM (3)} & \textbf{DMCM (4)} \\
\midrule
\textbf{Train Time (400 meta-grad)} & 679 & 428 & 350 & 331 & \textbf{317} \\
\textbf{Adapt Time (300 sine)} & 15.30 & \textbf{4.01} & 12.04 & 18.38 & 25.20 \\
\bottomrule
\end{tabular}
\caption{Training and adaptation time (seconds) at RTX 3080Ti GPU with 11th Gen Intel(R) Core(TM) i9-11900K @ 3.50GHz CPU.}
\vspace{-1.6em}
\label{tab:training_adaptation_time}
\end{table}

\subsection{Number of Parameter Variation at Context Vector}

\begin{figure}[h!]
  \centering
  \includegraphics[width=0.5\linewidth]{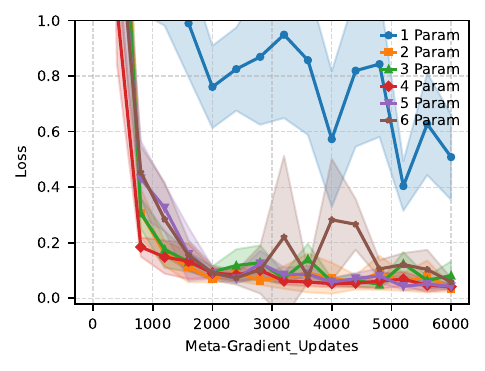}
  \caption{Training curves for the 2-context model (one entangled amplitude–phase context and one y-shift context) with different numbers of parameters for one context vector. Shaded regions indicate the 95\% confidence interval.}
  \label{fig:parameter}
\end{figure}
We evaluate the effect of varying the number of parameters in each context vector using a 2-context model (one entangled amplitude–phase context and one y-shift context). As shown in Fig.~\ref{fig:parameter}, a context vector with only a single parameter fails to capture the task variations, as expected, since the entangled context must simultaneously encode amplitude and phase shift, requiring at least two parameters. On the other hand, the relatively unstable learning and large deviation observed in the 6-parameter case suggest that overly large context vectors may reduce training stability and hinder generalization. Overall, these results highlight the importance of selecting an appropriate parameter dimension to balance expressiveness and stability.

\section{Go1 Dynamics Model}
\label{sec:Go1Dynamics}

\subsection{Hyperparameter and Experiment Settings}

For the Go1 dynamics modeling task, we train DMCM and CAVIA under closely aligned hyperparameters to ensure a fair comparison. Both models use an inner-loop learning rate of 0.1, a meta-learning rate of 0.001, and a fully connected network with four hidden layers of 512, 256, 128, 64 units, each with ReLU activations. The meta-batch size is set to 25 tasks per update, and each model employs a total of 40 context parameters (20 per context vector). We apply a warm-up of $B = 10$ iterations before meta-updates and perform $S_{\text{adapt}} = 10$ inner-loop updates for adaptation. The recombination loop is also applied.

\textbf{Data Collection and Prediction Target.}
We collected training data using the Isaac Gym simulator~\cite{makoviychuk2021isaac} and a naive walking policy explained at section~\ref{sec:naive}. The model learns to predict 33-dimensional outputs, including joint positions ($q$), joint velocities ($\dot{q}$), projected gravity, linear velocity, and angular velocity, from 45-dimensional input. The input consists of the robot's state and desired joint position ($q_{des}$) from the timestep 0.02 seconds earlier.

\textbf{Terrain Variation.}
We train across 44 types of terrains, including flat, wavy, sloped, and stair (both ascending and descending) terrains, as well as randomly structured terrains. Terrain diversity is further increased by randomizing friction coefficients and difficulty levels.

\textbf{Robot Property Variation.}
Robot-specific variations are introduced through domain randomization, summarized in Table~\ref{tab:robot_randomization2}. The randomized parameters include mass, center of mass (COM), action delay, control gains, and torque limits. Each property is perturbed independently across training tasks to enhance robustness.

\begin{table}[h!]
\centering
\small

\begin{tabular}{lcc}
\toprule
\textbf{Property} & \textbf{Min} & \textbf{Max} \\
\midrule
Mass offset (kg)             & $-1.0$     & $+4.0$  \\
Center of Mass shift (m)     & $-0.05$    & $+0.05$ \\
Action delay (s)                    & $0.0$      & $0.02$  \\
$K_p$ gain offset ($\text{N}\cdot\text{m}$)       & $-4.5$     & $+4.5$  \\
$K_d$ gain offset ($\text{N}{\cdot}\text{m}{\cdot}\text{s/rad}$)       & $-0.2$     & $+0.2$  \\
Torque limit offset ($\text{N}\cdot\text{m}$)     & $-4.0$     & $+4.0$  \\
\bottomrule
\end{tabular}
\caption{Randomized robot-specific properties and their value ranges used during training.}
\label{tab:robot_randomization2}
\end{table}

Default $K_p$ gain is set 20$\text{N}\cdot\text{m}$, default $K_d$ gain is set 0.5$\text{N}{\cdot}\text{m}{\cdot}\text{s/rad}$, default torque limit is set 30$\text{N}\cdot\text{m}$.

\textbf{Task Composition.}
We construct a dataset of $44 \times 350 = 15{,}400$ tasks, each corresponding to a unique combination of terrain and robot-specific properties. Each task contains 2,000 samples, equivalent to 40 seconds of walking. For inner-loop training, outer-loop updates, and evaluation, we randomly sample 700 data points per task.

\begin{figure}[b]
    \centering

    \begin{minipage}[t]{0.48\textwidth}
        \centering
        \includegraphics[width=\linewidth]{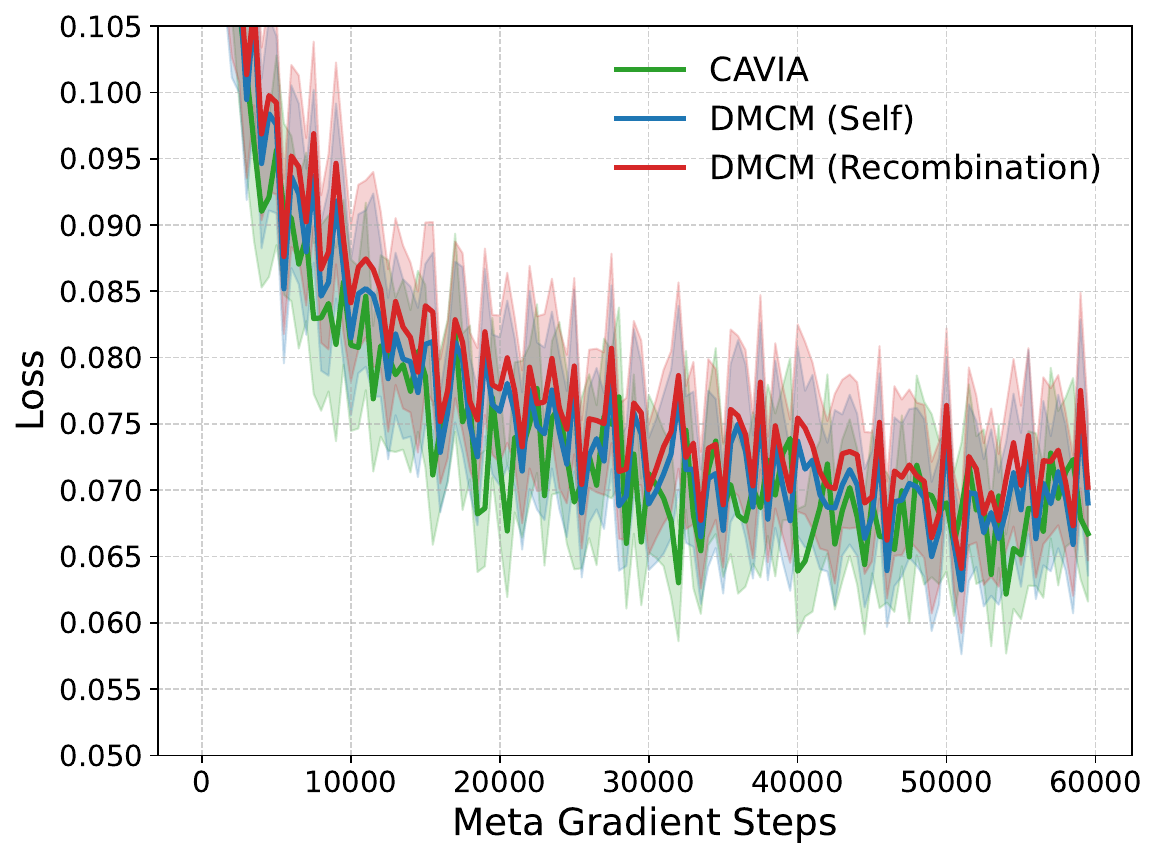}
        \caption{Dynamics model loss for CAVIA, DMCM with self-adaptation, and DMCM with recombination (zero-shot with shared context vectors). Shaded region indicates 95\% confidence interval. Evaluation is performed on 500 tasks.}
        \label{fig:DynamicsLoss}
    \end{minipage}
    \hfill
    \begin{minipage}[t]{0.48\textwidth}
        \centering
        \includegraphics[width=\linewidth]{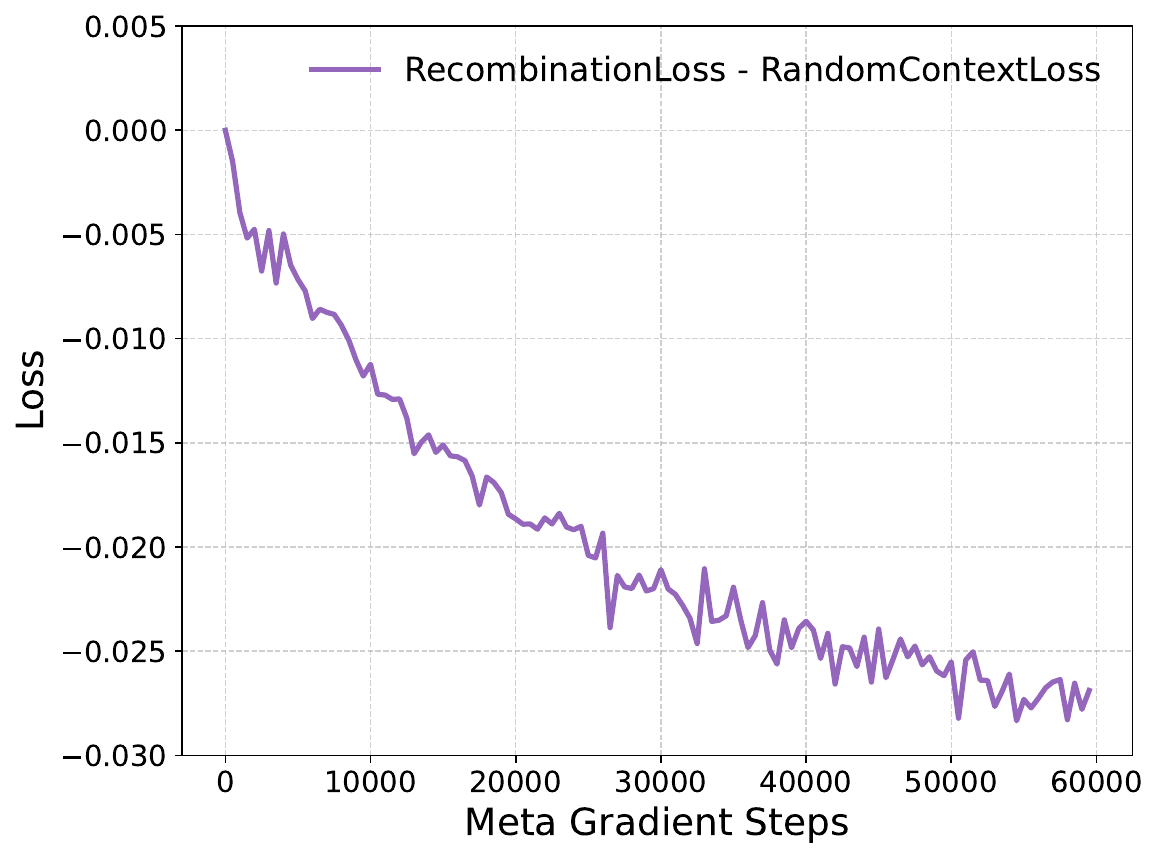}
        \caption{Loss gap between proper context sharing and randomly assigned context at DMCM. Negative values imply meaningful adaptation.}
        \label{fig:StupidSepar}
    \end{minipage}

\end{figure}

\label{sec:dynamicsHyper}

\subsection{Dynamics Model Training}
We train both CAVIA and DMCM dynamics models using the same dataset and aligned hyperparameters. All inputs and outputs are normalized by subtracting the mean and dividing by the standard deviation computed from the total training set. As shown in Fig.~\ref{fig:DynamicsLoss}, both models achieve comparable performance. Additionally, DMCM demonstrates reasonable performance under zero-shot conditions (recombination loss), using shared context vectors without access to task-specific data.

As shown in Fig.~\ref{fig:DynamicsLoss}, the DMCM model exhibits a steadily decreasing loss, indicating improved understanding of dynamics over the course of training. However, to ensure that the model truly benefits from task-specific adaptation, we must verify that it is not simply relying on shared parameters ($\theta$) that happen to generalize across all tasks. To assess this, we evaluate the model using randomly assigned context vectors that are not aligned with the target task. Fig.~\ref{fig:StupidSepar} shows the gap between recombination loss and random-context loss increases during DMCM training, indicating that the model is learning to encode useful, task-specific information.

\label{sec:dynamicsTraining}

\subsection{Additional Result:Sim Real Difference}
To evaluate DMCM’s ability to capture sim-to-real discrepancies, we conduct an additional experiment comparing simulated and real-world dynamics data. Real-world trajectories are collected on both flat and stair terrains using the Go1 robot, while simulation data are generated on corresponding flat and stair terrains using RaiSim~\cite{raisim}.

For a controlled comparison, we do not apply domain randomization, and robot-specific properties are matched as closely as possible between simulation and real-world conditions. This setup allows us to directly assess the sensitivity of the learned dynamics models to the domain gap that arises from sim-to-real transfer.
\begin{figure}[h!]
    \centering
    % --- First row ---
    \begin{subfigure}[t]{0.49\textwidth}
        \centering
        \includegraphics[width=\linewidth]{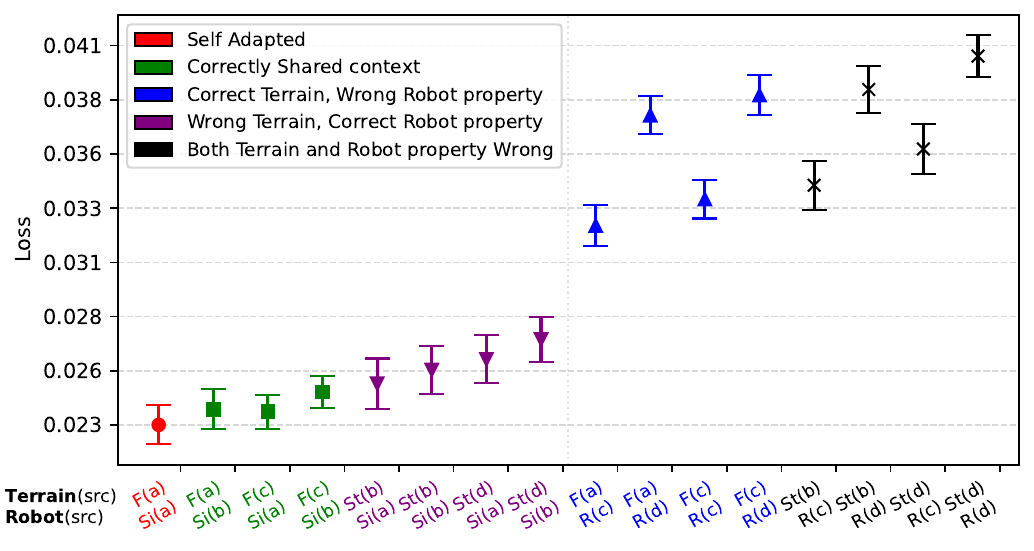}
        \caption{Flat, Sim}
        \label{fig:FlatSim}
    \end{subfigure}
    \hfill
    \begin{subfigure}[t]{0.49\textwidth}
        \centering
        \includegraphics[width=\linewidth]{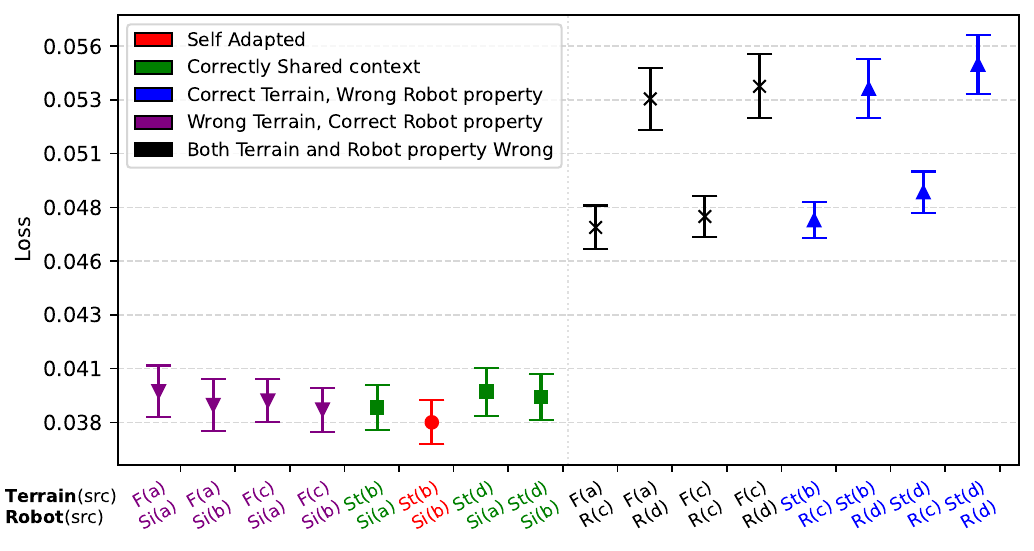}
        \caption{Stair, Sim}
        \label{fig:StairSim}
    \end{subfigure}
    
    % \vspace{0.8em} % <-- vertical gap between rows

    % --- Second row ---
    \begin{subfigure}[t]{0.49\textwidth}
        \centering
        \includegraphics[width=\linewidth]{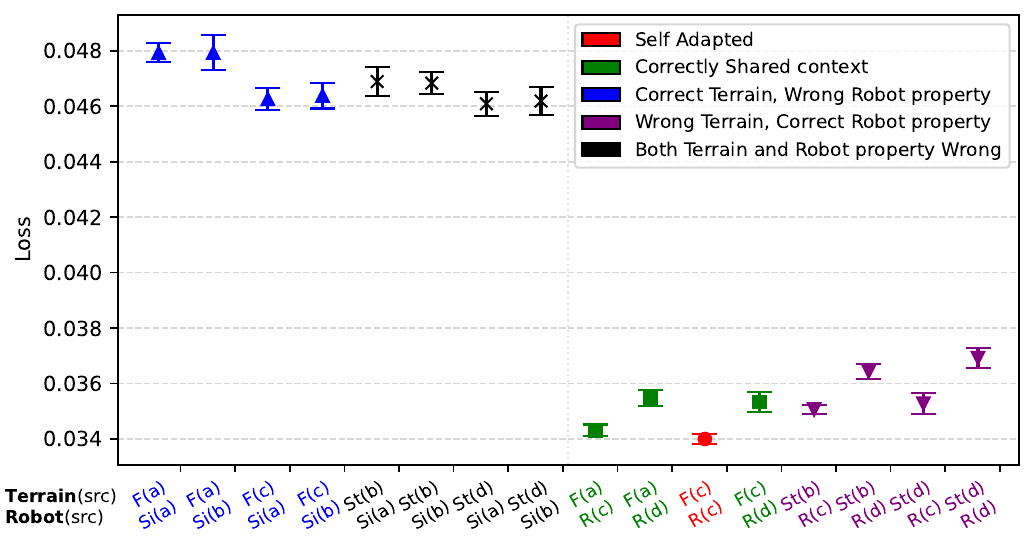}
        \caption{Flat, Real}
        \label{fig:FlatReal}
    \end{subfigure}
    \hfill
    \begin{subfigure}[t]{0.49\textwidth}
        \centering
        \includegraphics[width=\linewidth]{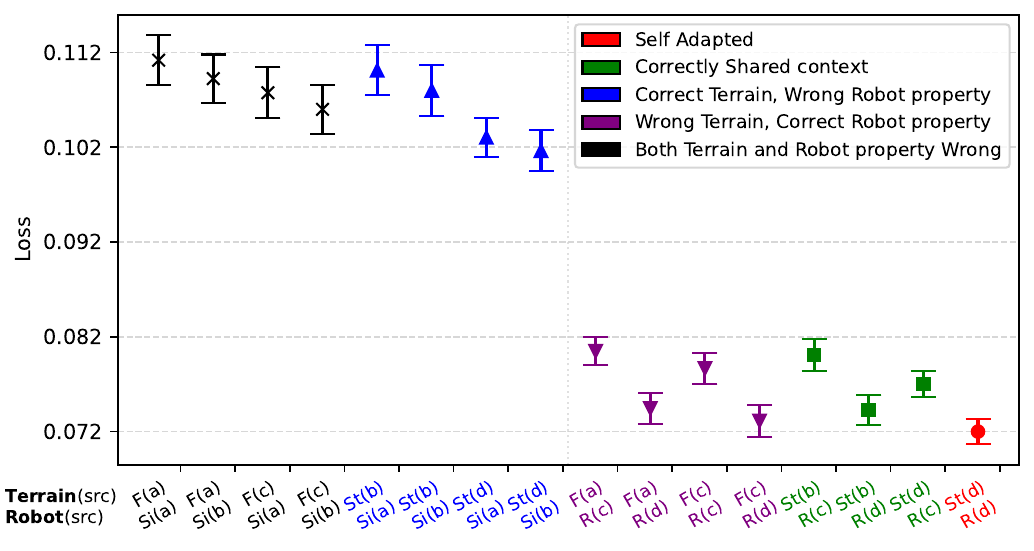}
        \caption{Stair, Real}
        \label{fig:StairReal}
    \end{subfigure}

    \caption{Each subplot shows the average loss for each dataset using 16 context vector sets derived from four
datasets. Each set includes a terrain context (Flat (F) or Stair (St)) and a robot property context (Sim (Si) or Real (R)). Results are averaged over 10 trials, with adaptation and test sets randomly sampled from 20 seconds within the 40 seconds of data. Error bars indicate the standard deviation.}
    \label{fig:simReal}
\end{figure}

As shown in Fig.~\ref{fig:simReal}, the DMCM model exhibits a noticeably higher loss when robot-specific properties differ between simulation and reality, compared to when only the terrain differs. This is observed even though the Raisim stair and the real stair differ in height. While this analysis does not fully explain the entire sim-to-real gap, the model’s predictions clearly reveal a measurable discrepancy in dynamics between the two domains, underscoring DMCM’s ability to capture variations that arise from differences in robot embodiment.

\section{Go1 Reinforcement Learning Policy}
\subsection{Training Hyperparameter of the Reinforcement Learning Policy}

To train the Reinforcement Learning(RL) policy of quadrupedal robot locomotion for rough terrain, we employ PPO \cite{schulman2017proximal} as our policy gradient algorithm within the Isaac Gym simulator \cite{makoviychuk2021isaac}.

We use a concurrent training architecture to estimate proprioceptive value while learning the policy \cite{ji2022concurrent}.

\subsubsection{Naive Policy}
For the training curriculum we use only pyramid stair and slope terrain from \cite{rudin2022learning} with 4096 parallel environments. We largely follow the motor gains, reward functions and randomization settings  from \cite{margolis2023walk} to enable robust sim-to-real transfer. Detailed input and hyperparameters are in Table~\ref{S_Tab:observation}, \ref{S_Tab:hyperparameters_ppo}.

\begin{table}[h!]
\centering
\begin{minipage}[t]{0.48\textwidth}
\centering
\resizebox{\textwidth}{!}{
\begin{tabular}{@{}lll@{}}
\toprule
\textbf{Observation Type} & \textbf{Input} & \textbf{Dim.} \\
\midrule
\multirow{10}{*}{\textbf{Proprioception}}    
& Linear body velocity estimation       & 3  \\
& Angular body velocity                 & 3  \\
& Body height estimation                & 1  \\
& Foot height estimation                & 4  \\
& Contact probability estimation        & 4  \\
& Command                               & 3  \\
& Projected gravity vector              & 3  \\
& Action                                & 12 \\
& Joint position                        & 12 \\
& Joint velocity                        & 12 \\
& Action (2 time steps ago)            & 12 \\
& Joint position (2 time steps ago)    & 12 \\
& Joint velocity history (2 steps ago) & 12 \\
& Action (4 time steps ago)            & 12 \\
& Joint position (4 time steps ago)    & 12 \\
& Joint velocity history (4 steps ago) & 12 \\
\bottomrule
\end{tabular}
}
\caption{Observation Types and Dimensions}
\label{S_Tab:observation}
\end{minipage}%
\hfill
\begin{minipage}[t]{0.48\textwidth}
\centering
\begin{tabular}{@{}ll@{}}
\toprule
\textbf{Parameter}         & \textbf{Value} \\
\midrule
horizon length (dt: 0.02)  & 25       \\
learning rate              & 3.0E-4    \\
kl threshold               & 0.008    \\
discount factor            & 0.99     \\
entropy coef               & 0.001    \\
clip ratio                 & 0.2      \\
batch size                 & 102400   \\
mini batch size            & 20480    \\
\bottomrule
\end{tabular}
\caption{Hyperparameters for PPO, naive policy}
\label{S_Tab:hyperparameters_ppo}
\end{minipage}
\end{table}

For the linear body velocity, body height, foot height and contact probability, we use the concurrent estimation strategy using ground truth data from simulation. We also note that we give true value to the critic network during training the policy. 

\subsubsection{Remaining Policies}

Beyond the naive policy environments, we expand the training curriculum by incorporating wavy and randomly distributed terrains and by increasing the difficulty levels of the pyramid stair and slope terrains. These enhanced environments are trained using 8,816 parallel simulation instances.

The input observations remain consistent with those listed in Table~\ref{S_Tab:observation}, while context parameters are concatenated at single-CAVIA policy and context vectors are concatenated at multi-DMCM policy. Training is performed using the hyperparameters detailed in Table~\ref{S_Tab:hyperparameters_ppo_new}.

\begin{table}[h!]
\centering

\begin{tabular}{@{}ll@{}}
\toprule
\textbf{Parameter}                 & \textbf{Value} \\ \midrule
horizon length (dt : 0.02)         & 25        \\
learning rate                      & 3.0 E-4        \\
kl threshold          & 0.008         \\
discount factor                    & 0.99          \\
entropy coef                     & 0.001              \\
clip ratio                         & 0.2            \\
batch size                         & 220400           \\
mini batch size                         & 44080           \\ \bottomrule
\end{tabular}

\caption{Hyperparameters for PPO, rest of the policies}
\label{S_Tab:hyperparameters_ppo_new}

\end{table}

\subsection{Data Acquisition with Naive Policy}
\label{sec:naive}

For the dynamics modeling and context collecting for reinforcement learning experiments, a naive policy is employed for data acquisition. This policy performs significantly worse than any of the trained policies, including the vanilla policy, as shown in Table~\ref{tab:naive_vs_vanila}. The evaluation is conducted under the same experimental conditions described in Appendix~\ref{sec:eval_condition}. The naive policy is selected to ensure that context learning is not biased by high-performing behavior and that performance improvements in the trained policies are not attributable to favorable data collection.

%Can be changed(space)
\vspace{2em}

\begin{table}[h]
\centering
\begin{tabular}{lcc}
\hline
\textbf{Metric} & \textbf{Naive} & \textbf{Vanilla} \\
\hline
\multicolumn{3}{c}{\textbf{Inner Distribution}} \\
\hline
Success Counter & 404/2000 & \textbf{1580/2000} \\
RMSE Linear Vel (m/s) & 0.2276 & \textbf{0.1552} \\
Reward (w/o Termination Reward) & 12.0252 & \textbf{17.4848} \\
Lifespan & 0.5886 & \textbf{0.9190} \\

AvgDist (Success) (m) & 4.8803 & \textbf{7.3806} \\
\hline
\multicolumn{3}{c}{\textbf{OOD Terrain}} \\
\hline
Success Counter & 98/2000 & \textbf{205/2000} \\
RMSE Linear Vel (m/s) & 0.2912 & \textbf{0.2555} \\
Reward (w/o Termination Reward) & 5.9131 & \textbf{8.5807} \\
Lifespan & 0.3522 & \textbf{0.4737} \\
% AvgDist (All) & \textbf{3.3568} & 3.3028 \\
% AvgDist (Success) & 3.6103 & \textbf{3.8804} \\
\hline
\multicolumn{3}{c}{\textbf{OOD Robot-Specific Properties}} \\
\hline
Success Counter & 0/2000 & \textbf{48/2000} \\
Lifespan & 0.2586 & \textbf{0.3050} \\

\hline
\end{tabular}
\caption{Comparison between Naive and Vanilla policies across inner distribution, OOD terrain, and OOD robot-specific properties. }
\label{tab:naive_vs_vanila}
\end{table}

Context data for both the multi-DMCM and single-CAVIA policies are collected using this naive policy. The dataset comprises trajectories from 58 different terrains and 500 randomly sampled robot-specific property configurations (Table~\ref{tab:rl_randomization3}), yielding $58 \times 500 = 29{,}000$ tasks, each consisting of 2000 timesteps (40 seconds) of walking data.

\begin{table}[h!]
\centering
\small

\begin{tabular}{lcc}
\toprule
\textbf{Property} & \textbf{Min} & \textbf{Max} \\
\midrule
Mass offset (kg)             & $-1.0$     & $+4.0$  \\
Center of Mass shift (m)     & $-0.05$    & $+0.05$ \\
Action delay (s)            & $0.0$      & $0.02$  \\
$K_p$ gain offset ($\text{N}\cdot\text{m}$)       & $-2.5$     & $+2.5$  \\
$K_d$ gain offset ($\text{N}{\cdot}\text{m}{\cdot}\text{s/rad}$)       & $-0.2$     & $+0.2$  \\
\bottomrule
\end{tabular}
\caption{Randomized robot-specific properties and their value ranges used during RL training.}
\label{tab:rl_randomization3}
\end{table}

\vspace{-1.0em}
Compared to Table~\ref{tab:robot_randomization2}, torque limit variation is removed and the $K_p$ gain range is slightly reduced. For each task, we randomly sample 700 points three times to generate context parameters and context vectors. This produces three context parameter sets per task for the single-CAVIA model, and three terrain context vectors along with three robot-specific property context vectors per task for the multi-DMCM model.

\textbf{Real-World Data Collection:} The same naive policy is used for real-world data collection. The following seven conditions are used to represent diverse robot-terrain settings:
\begin{enumerate}
\item Flat terrain
\item Flat terrain with 1.5kg payload
\item Climbing stairs (17cm depth)
\item Wavy terrain
\item Wavy terrain with 1.5kg payload
\item Flat terrain with 1.5kg payload and water bottle on legs
\item Flat terrain with 16$\text{N}\cdot\text{m}$ $K_p$ (low gain) and 1.5kg payload
\end{enumerate}
Data collection on stairs with a 1.5 kg payload was not possible, as the naive policy was unable to successfully complete the task. These datasets are used for adapting the CAVIA and DMCM dynamics models. The adapted contexts are then applied in the single-CAVIA and multi-DMCM policies for real-world deployment.

\subsection{Training and Experiment Settings}

During training, the context parameters or vectors are concatenated with the corresponding observations.
For the single-CAVIA policy, one of the three generated context parameter sets is randomly selected at the beginning of each episode.
In contrast, the disentangled structure of DMCM allows the multi-DMCM policy to reuse context vectors derived from the same factor of variation across different tasks.
During training, at multi-DMCM policy, randomly assigned context vectors that share the same underlying context are concatenated with observations. Additional analysis is provided in Appendix~\ref{sec:multi_context_compar}.

\textbf{Terrain Setup:} We divide the environment terrains into the following 8 types:
\begin{enumerate}
\item Low stairs up
\item Low stairs down
\item Wavy terrain
\item Sloped terrain
\item Random terrain
\item High stairs down (with velocity limits)
\item High stairs up (with velocity limits)
\item High stairs up (with velocity limits and low friction)

\end{enumerate}

\textbf{Velocity Limits:} For high stairs, the linear command x directional velocity ranges from $0.4\,\mathrm{m/s}$ to $0.6\,\mathrm{m/s}$, y directional velocity as 0, and yaw rate from $-0.6\,\mathrm{rad/s}$ to $0.6\,\mathrm{rad/s}$.
Except for high stairs, the linear command x directional velocity ranges from $-1.0\,\mathrm{m/s}$ to $1.0\,\mathrm{m/s}$, y directional velocity from $-1.0\,\mathrm{m/s}$ to $1.0\,\mathrm{m/s}$, and yaw rate from $-\pi\,\mathrm{rad/s}$ to $\pi\,\mathrm{rad/s}$.

\textbf{Robot-Specific Properties:} Each environment is assigned one configuration sampled from Table~\ref{tab:rl_randomization3}.

\textbf{Policy Selection:} For all policies (vanilla, single-CAVIA, multi-DMCM), trained policy is selected with the highest reward after training over 33,000 episodes.

In this experiment, we focus on developing a policy that can climb stairs effectively. Therefore, 40\% of environments are assigned to stair climbing task. 

\label{sec:RLenvSetting}

\subsection{Evaluation Settings}

\begin{figure}[h!]
\centering
\begin{subfigure}[t]{0.48\textwidth}
\centering
\includegraphics[width=\linewidth]{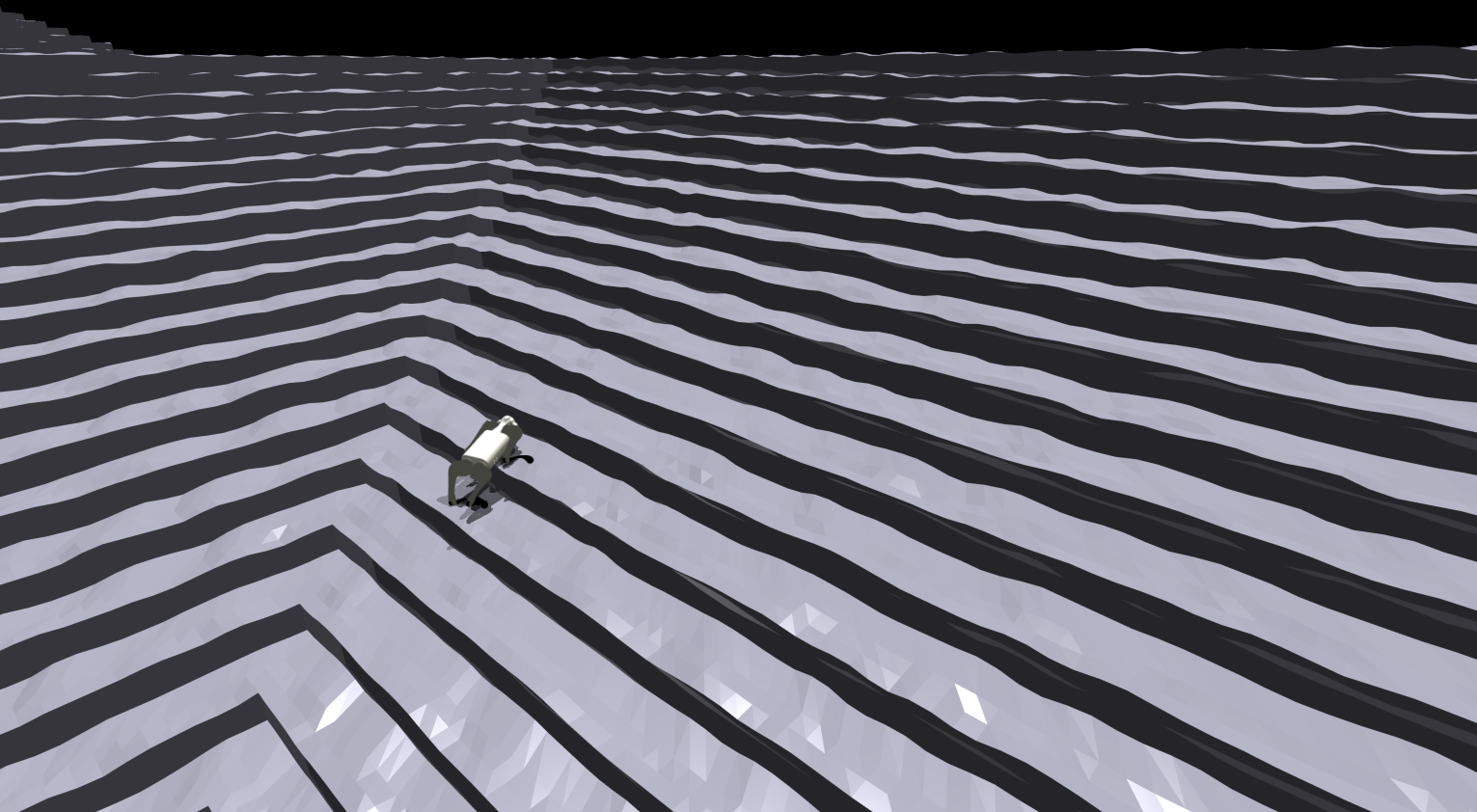}
\caption{Highest level stair terrain}
\label{fig:highstair}
\end{subfigure}
\hfill
\begin{subfigure}[t]{0.48\textwidth}
\centering
\includegraphics[width=\linewidth]{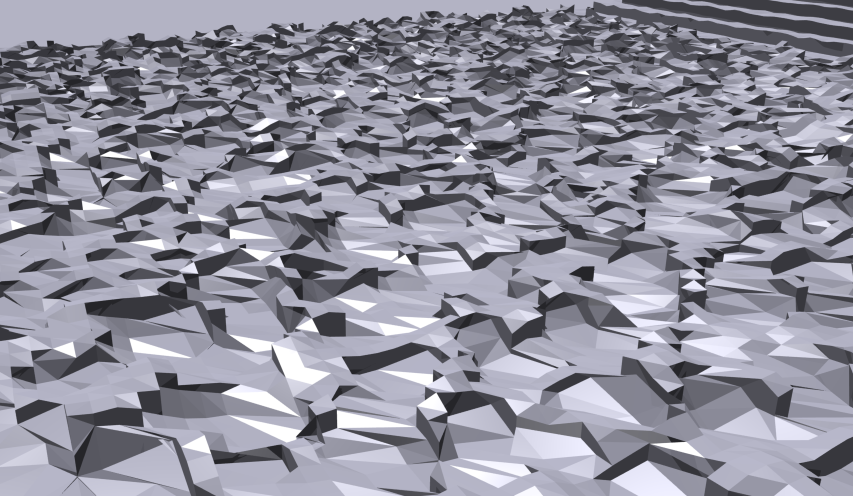}
\caption{Difficult Randomly distributed terrain used for OOD terrain}
\label{fig:random}
\end{subfigure}
\caption{Terrains used for evaluation
}
\label{fig:terraineval}
\end{figure}

Three environments are primarily used for evaluation in this study. The first is a high-stair environment, one of the most challenging terrains in the training settings, with a stair height of 0.27 m and a width of 0.31 m. The second is a highly randomized out-of-distribution (OOD) terrain, illustrated in Fig.~\ref{fig:random}. For both environments, the same robot-specific property distributions as in training are applied. The third evaluation case involves OOD robot-specific properties, as listed in Table~\ref{tab:rl_randomization_eval}, while using the same terrain as the first environment, the steep stair.

\vspace{3em}

\begin{table}[h!]
\centering
\small

\begin{tabular}{lcc}
\toprule
\textbf{Property} & \textbf{Min} & \textbf{Max} \\
\midrule
Mass offset (kg)             & $5.0$     & $6.0$  \\
Center of Mass shift (m)     & $-0.05$    & $+0.05$ \\
Action delay (s)            & $0.0$      & $0.02$  \\
$K_p$ gain offset ($\text{N}\cdot\text{m}$)       & $-5.0$     & $-4.0$  \\
$K_d$ gain offset ($\text{N}{\cdot}\text{m}{\cdot}\text{s/rad}$)       & $-0.2$     & $+0.2$  \\
\bottomrule
\end{tabular}
\caption{Randomized robot-specific properties and their value ranges used in the OOD robot-specific evaluation.}
\label{tab:rl_randomization_eval}
\end{table}

\label{sec:eval_condition}

\subsection{Robustness via Context Sharing at Multi-DMCM Policy }
In the single-CAVIA setup, each task is associated with three distinct context parameter sets, which are restricted to that specific task and cannot be reused elsewhere.
In contrast, DMCM’s disentangled structure allows the multi-DMCM policy to share partial contexts across different tasks. For instance, a robot-specific context learned from one task can be combined with a terrain context learned from another.
We further investigate whether evaluation performance improves when probabilistically assigning context vectors that share the same underlying factor of variation, or deterministically using the context vectors derived under the exact same conditions.

\begin{table}[h!]
\centering
\begin{tabular}{lcc}
\hline
\textbf{Metric} & \textbf{Multi-DMCM (Probabilistic)} & \textbf{Multi-DMCM (Deterministic)} \\
\hline
\multicolumn{3}{c}{\textbf{Inner Distribution}} \\
\hline
Success Counter & \textbf{1892/2000} & 1688/2000 \\
RMSE Linear Vel (m/s) & \textbf{0.0816} & 0.0992 \\
Reward (w/o Termination Reward) & \textbf{21.9682} & 19.4695 \\
Lifespan & \textbf{0.9851} & 0.9433 \\
% AvgDist (All) & \textbf{7.8833} & 6.4964 \\
AvgDist (Success) (m) & \textbf{8.4581} & 8.0676 \\
\hline
\multicolumn{3}{c}{\textbf{OOD Terrain}} \\
\hline
Success Counter & 281/2000 & \textbf{431/2000} \\
RMSE Linear Vel (m/s) & \textbf{0.2486} & 0.2607 \\
Reward (w/o Termination Reward) & \textbf{10.0295} & 9.1101 \\
Lifespan & 0.5476 & \textbf{0.6104} \\
% AvgDist (All) & \textbf{3.5764} & 3.7341 \\
% AvgDist (Success) & \textbf{4.1294} & 3.8165 \\
\hline
\multicolumn{3}{c}{\textbf{OOD Robot-Specific Properties }} \\
\hline
Success Counter & \textbf{602/2000} & 151/2000 \\
RMSE Linear Vel (m/s) & \textbf{0.0944} & 0.1083 \\
Reward (w/o Termination Reward) & \textbf{19.2283} & 17.7090 \\
Lifespan & \textbf{0.6009} & 0.3813 \\
% AvgDist (All) & \textbf{2.9941} & 2.7767 \\
AvgDist (Success) (m) & \textbf{7.9205} & 7.2410 \\
\hline
\end{tabular}
\caption{Performance comparison of multi-DMCM (Probabilistic) and multi-DMCM (Deterministic) across inner distribution, OOD terrain, and OOD robot-specific properties. Best values in each row are bolded.}
\label{tab:multi_random_vs_deterministic_full}
\end{table}
The results in Table~\ref{tab:multi_random_vs_deterministic_full} demonstrate that the multi-DMCM policy generally performs better when using context vectors probabilistically selected from other tasks sharing the same factor, rather than those derived from the current task. This property suggests potential benefits for real-world deployment, as it reflects the policy’s robustness in interpreting context vectors. For the multi-DMCM policy, a probabilistically selected context configuration is used during all the other evaluations.

\label{sec:multi_context_compar}

\subsection{Verifying Policy's Context Utilization}
\label{sec:verify}

\subsubsection{Policy Performance with Incorrect Context}
We evaluate the performance of both single-CAVIA and multi-DMCM policies. However, their performance may not degrade significantly even when incorrect context vectors or parameters are used, raising questions about whether they truly utilize context information. Therefore, we check the performance with context parameters and vectors derived from the wrong terrain context and randomly selected robot-specific contexts on the steep stair evaluation (Fig.\ref{fig:highstair}). Specifically, we assign the terrain context of stair down instead of stair up. 

\begin{table}[h!]
\centering
\begin{tabular}{lccccc}
\toprule
\textbf{Metric} & \textbf{Multi-DMCM} & \textbf{Multi (Wrong)} & \textbf{Single-CAVIA} & \textbf{Single (Wrong)} & \textbf{Vanilla} \\
\midrule
Success & 1892/2000 & 1323/2000 & 1905/2000 & 1339/2000 & 1580/2000 \\
RMSE Lin Vel (m/s) & 0.0816 & 0.0931 & 0.0798 & 0.1232 & 0.1552 \\
Reward & 21.97 & 20.32 & 22.18 & 18.57 & 17.48 \\
Lifespan & 0.9851 & 0.8652 & 0.9852 & 0.8598 & 0.9190 \\

AvgDist (Success) (m) & 8.4581 & 8.3407 & 8.5203 & 7.3069 & 7.3806 \\
\bottomrule
\end{tabular}
\caption{Comparison across different methods under inner distribution conditions.}
\label{tab:inner_comparison_transposed}
\end{table}
% \vspace{-1.0em}

As Table~\ref{tab:inner_comparison_transposed} shows, performance drops across all metrics for both policies when incorrect context information is used, with success rate and lifespan falling below those of the vanilla policy. These results confirm that both single-CAVIA and multi-DMCM policies actively rely on context information during deployment.

\subsubsection{Effect of Contexts on Multi-DMCM Policy}
To examine whether terrain and robot-specific property contexts meaningfully influence behavior in the multi-DMCM policy, we design controlled experiments that isolate the effect of each context type. Two types of variations are considered:  
(i) \textbf{Robot-specific context variation}: payload differences.  
(ii) \textbf{Terrain context variation}: changes in stair level and type (ascending and descending).  

The robot is evaluated on flat terrain without payload as the baseline. Then, incorrect context vectors are deliberately assigned from variation cases (i) and (ii). For the robot-specific context variation, assigning a high-payload context to the no-payload setting increased the robot’s body height (Table~\ref{tab:mean_body_height}), suggesting that the policy compensates for weight that is not actually present. For the terrain context variation, assigning a steep-terrain context on flat ground increased the standard deviation of the front foot’s height (Table~\ref{tab:front_feet_std}), indicating that the blind policy adapts by probing more aggressively with its front legs to detect terrain changes, prioritizing safety at the cost of stability.

% \vspace{-0.8em}
\begin{table}[h!]
\centering
\begin{tabular}{lccccc}
\toprule
\textbf{Mass for Context} & \textbf{-1.5kg} & \textbf{0kg} & \textbf{3kg} & \textbf{6kg} & \textbf{9kg} \\
\midrule
\textbf{Mean Body Height (m)} & 0.327 & 0.329 & 0.341 & 0.374 & 0.387 \\
\bottomrule
\end{tabular}
% \vspace{-0.2em}
\caption{Mean body height under different mass contexts.}
\label{tab:mean_body_height}
\end{table}
\vspace{-1.0em}

% \vspace{-1.6em}
\begin{table}[h!]
\centering
% \scriptsize
% \setlength{\tabcolsep}{1.5pt} % tighter column spacing
% \renewcommand{\arraystretch}{0.9} % slightly reduce row height
\begin{tabular}{lccccc}
\toprule
\textbf{Terrain} & High Down & Low Down & Flat & Low Up & High Up \\
\midrule
\textbf{Std (L/R)} & 0.0283 / 0.0351 & 0.0236 / 0.0329 & 0.0215 / 0.0316 & 0.0240 / 0.0333 & 0.0271 / 0.0352 \\
\bottomrule
\end{tabular}
% \vspace{-0.52em}
\caption{ Standard deviation of front leg heights (Left/Right) under different terrain contexts.}
\label{tab:front_feet_std}
\end{table}
% \vspace{-1.0em}

\subsection{Additional Real-World Experiments}
We further evaluate the performance of the multi-DMCM, single-CAVIA, and vanilla policies on a 17\,cm stair-climbing task, followed by more challenging scenarios designed for the multi-DMCM model.

\label{sec:AddReal}

\subsubsection{Stair with 1.5kg Payload}
All three policies successfully complete the 17 cm stair-climbing task with a 1.5 kg payload.
For the multi-DMCM policy, the terrain context is taken from Isaac Gym simulation data of an 18 cm stair climb, while the robot-specific context comes from real-world flat-terrain data with a 1.5 kg payload.
The single-CAVIA policy uses Isaac Gym data of an 18 cm stair climb with a 1.5 kg payload, with other robot-specific properties without domain randomization. Despite the added mass, all models complete the task without failure. 

\subsubsection{Multi-DMCM Policy in Complex Scenarios}
\begin{figure}[h!]
    \centering
    \includegraphics[width=0.45\linewidth]{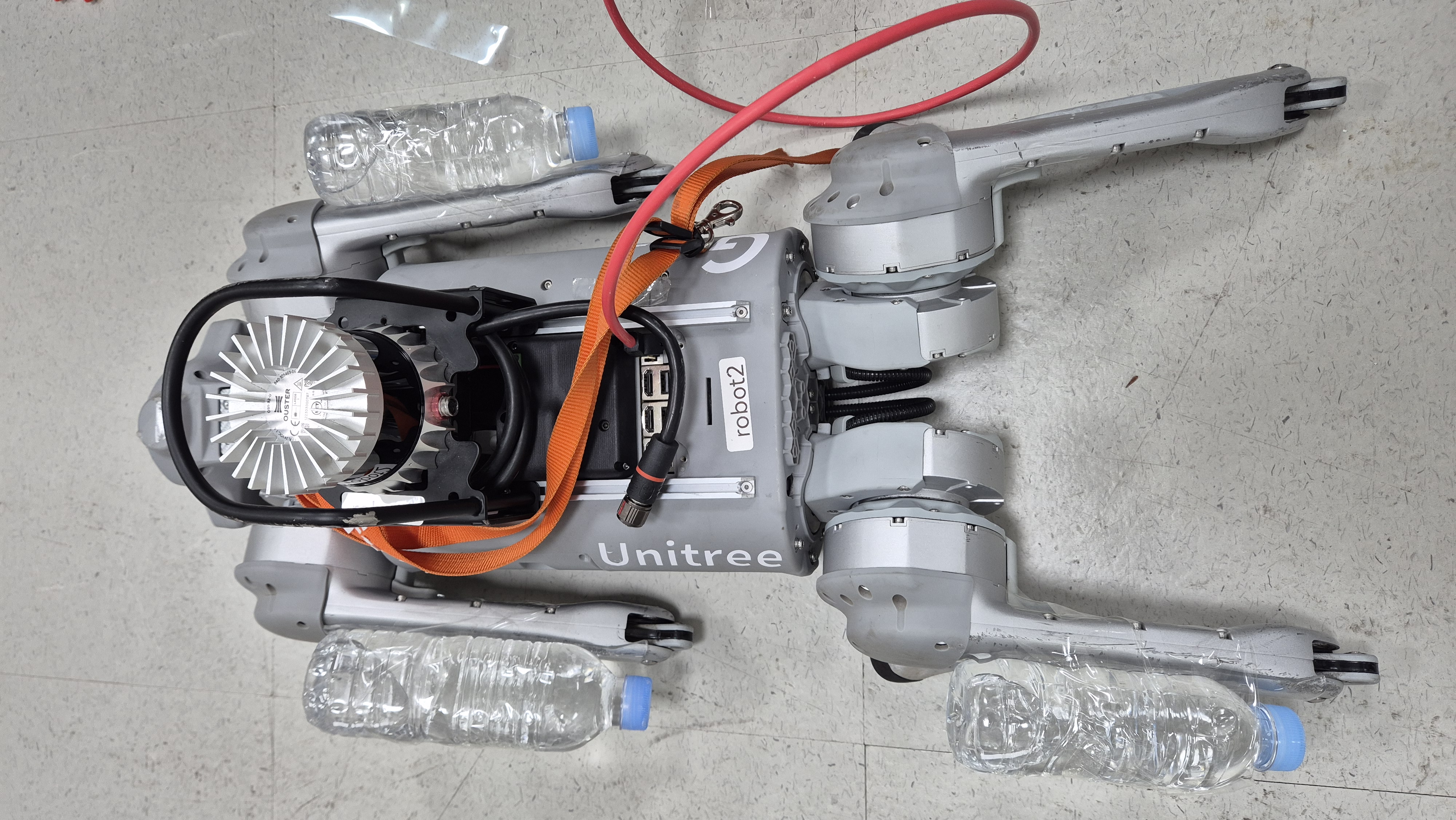}
    \caption{Go1 robot with water bottles attached at three legs}
    \label{fig:bottle}
\end{figure}

To further test robustness under more complex conditions, asymmetrical payloads are added to the robot’s legs: a 500 ml water bottle on the front left leg, and 300 ml bottles on the front right and rear left legs (Fig.~\ref{fig:bottle}).

Using context vectors obtained from previously collected data, the multi-DMCM policy successfully navigates both stair and wavy terrains.
For the stair task, the terrain context is taken from simulation data of an 18cm stair climb, and the robot-specific context is taken from real-world flat-terrain data with the water-bottle payload.
For the wavy terrain, the terrain context comes from real-world wavy-terrain data without payload, while the flat-terrain water bottle setup context is reused for robot-specific context.
In both scenarios, the multi-DMCM policy achieves stable locomotion with no failures.

\clearpage

% The acknowledgments are automatically included only in the final and preprint versions of the paper.
% \acknowledgments{If a paper is accepted, the final camera-ready version will (and probably should) include acknowledgments. All acknowledgments go at the end of the paper, including thanks to reviewers who gave useful comments, to colleagues who contributed to the ideas, and to funding agencies and corporate sponsors that provided financial support.}

%===============================================================================

% no \bibliographystyle is required, since the corl style is automatically used.

\end{document}